\documentclass[journal,final,twoside]{IEEEtran}

\ifCLASSINFOpdf
  \usepackage[pdftex]{graphicx}
\else
 \usepackage[dvips]{graphicx}
\fi
\usepackage{algorithm}
\usepackage{algorithmic}
\usepackage{array}
\usepackage[colorlinks=true, linkcolor=black,anchorcolor=black,citecolor=black,urlcolor=black]{hyperref}
\usepackage{url}
\usepackage{amssymb}
\usepackage{amsmath}
\usepackage{amssymb}
\usepackage{multirow}
\usepackage{cite}
\usepackage{bm}
\usepackage{xcolor}
\usepackage{subfigure}     
\newtheorem{thm}{Theorem}

\newtheorem{definition}{Definition}
\newtheorem{lemma}{Lemma}

\begin{document}
\title{Graph Neural Networks With Lifting-based Adaptive Graph Wavelets}

\author{
Mingxing~Xu,
Wenrui~Dai,~\IEEEmembership{Member,~IEEE,}
Chenglin~Li,~\IEEEmembership{Member,~IEEE,}
Junni~Zou,~\IEEEmembership{Member,~IEEE,}
Hongkai~Xiong,~\IEEEmembership{Senior~Member,~IEEE,}
and~Pascal~Frossard,~\IEEEmembership{Fellow,~IEEE}
\thanks{This work was supported in part by the National Natural Science Foundation of China under Grants 61932022, 61720106001, 61931023, 61831018, 61971285, 61871267, 61972256, and 62125109, and in part by the Program of Shanghai Science and Technology Innovation Project under Grant 20511100100.  Mingxing~Xu was sponsored by China Scholarship Council (CSC) for one year's study at the \'{E}cole Polytechnique F\'{e}d\'{e}rale de Lausanne (EPFL).  (\emph{Corresponding author: Wenrui Dai}.)}
\thanks{M. Xu, C. Li, and H. Xiong are with the Department of Electronic Engineering, Shanghai Jiao Tong University, Shanghai 200240, China. E-mail: xumingxing@sjtu.edu.cn, lcl1985@sjtu.edu.cn, xionghongkai@sjtu.edu.cn.}
\thanks{W. Dai and Z. Jou are with the Department of Computer Science and Engineering, Shanghai Jiao Tong University, Shanghai 200240, China. E-mail:daiwenrui@sjtu.edu.cn, zoujunni@sjtu.edu.cn.}
\thanks{P. Frossard is with the Signal Processing Laboratory (LTS4), \'{E}cole Polytechnique F\'{e}d\'{e}rale de Lausanne (EPFL), CH-1015, Lausanne, Switzerland. E-mail: pascal.frossard@epfl.ch}
\thanks{This paper has supplementary downloadable material available at http://ieeexplore.ieee.org provided by the authors.}
}

\markboth{IEEE Transactions on Signal and Information Processing over Networks}
{Xu \MakeLowercase{\textit{et al.}}: Graph Neural Networks With Lifting-based Adaptive Graph Wavelets}

\maketitle

\begin{abstract}
Spectral-based graph neural networks (SGNNs) have been attracting increasing attention in graph representation learning. However, existing SGNNs are limited in implementing graph filters with rigid transforms (e.g., graph Fourier or predefined graph wavelet transforms) and cannot adapt to signals residing on graphs and tasks at hand. In this paper, we propose a novel class of graph neural networks that realizes graph filters with adaptive graph wavelets. Specifically, the adaptive graph wavelets are learned with neural network-parameterized lifting structures, where structure-aware attention-based lifting operations (i.e., prediction and update operations) are developed to jointly consider graph structures and node features. We propose to lift based on diffusion wavelets to alleviate the structural information loss induced by partitioning non-bipartite graphs. By design, the locality and sparsity of the resulting wavelet transform as well as the scalability of the lifting structure are guaranteed. We further derive a soft-thresholding filtering operation by learning sparse graph representations in terms of the learned wavelets, yielding a localized, efficient, and scalable wavelet-based graph filters. To ensure that the learned graph representations are invariant to node permutations, a layer is employed at the input of the networks to reorder the nodes according to their local topology information. We evaluate the proposed networks in both node-level and graph-level representation learning tasks on benchmark citation and bioinformatics graph datasets. Extensive experiments demonstrate the superiority of the proposed networks over existing SGNNs in terms of accuracy, efficiency, and scalability.
\end{abstract}
\section{Introduction}\label{sec:intro}
\IEEEPARstart{G}{raphs} are flexible and powerful data representation forms that could describe attributes/features of individual objects as well as the interactions among them. Examples include data defined on social networks, traffic networks, bioinformatics networks, and chemical compounds. Representation learning on graphs \cite{DBLP:journals/debu/HamiltonYL17,hamilton2020graph} aims to jointly encode the node attributes and graph structures as a compact low-dimensional vector to facilitate downstream machine learning tasks, e.g., node classification and graph classification. The unprecedented successes achieved by convolutional neural networks (CNNs) in  representation learning for Euclidean data have ignited the ever-increasing interest in generalizing CNNs to graph representation learning, aka graph neural networks (GNNs)~\cite{bronstein2017geometric,zhou2020graph,wu2020comprehensive,zhang2020deep}. In particular, spectral-based graph neural networks~\cite{defferrard2016convolutional,kipf2016semi,kipf2016variational,levie2018cayleynets,bianchi2021graph,asiri2019dfnets} grounded on spectral graph theory~\cite{chung1997spectral} and graph signal processing~\cite{ortega2018graph} have been attracting increasing attention due to their ability to explore graph signals from a graph spectral perspective and their interpretability via the established theory of graph signal processing~\cite{shuman2013emerging} (e.g., graph filter banks~\cite{tremblay2018design,gama2020stability}, scattering transform~\cite{bruna2013invariant,zou2020graph}, and frame theory~\cite{hammond2011wavelets}).

Existing spectral-based graph neural networks can be roughly classified into two categories according to the transform domain in which graph filters are realized, namely Fourier-based SGNNs~\cite{DBLP:journals/corr/HenaffBL15,defferrard2016convolutional,kipf2016semi,levie2018cayleynets} and wavelet-based SGNNs~\cite{xu2018graph,li2020fast,zheng2021how}. The Fourier-based SGNNs perform filtering in the graph Fourier domain~\cite{DBLP:journals/corr/HenaffBL15,defferrard2016convolutional}. The Fourier coefficients obtained with graph Fourier transform (GFT) are selectively amplified or attenuated with spectral filters before being transformed back to vertex domain through the inverse graph Fourier transform. However, on the one hand, the global Fourier transform limits the freedom of degree of graph filters in extracting expressive representations in terms of local features. On the other hand, the dense Fourier basis leads to heavy computation and memory overheads. Consequently, some constraints (e.g., smoothness) are placed on the learned graph filters for efficient and spatially localized graph feature extraction. For example, graph filters are usually restricted to be smooth in the spectral domain~\cite{DBLP:journals/corr/HenaffBL15} or in the spaces spanned by low-order polynomials (e.g., Chebyshev~\cite{defferrard2016convolutional} or rational polynomials~\cite{levie2018cayleynets}). Alternatively, wavelet-based SGNNs (graph wavelet networks for short) implement graph filters in the wavelet domain to leverage the inherent sparsity and locality of graph wavelet transforms~\cite{xu2018graph,li2020fast,zheng2021how}, which opens up a new way for the design of SGNNs. However, all these graph wavelet networks~\cite{xu2018graph,li2020fast,zheng2021how} construct graph filters with manually-designed graph wavelets (e.g., predefined with some wavelet generating functions with fixed spectral shapes) which can not adapt to graph signals and tasks at hand. Therefore, the capacity of graph wavelet networks is not fully exploited yet. Moreover, for each graph wavelet filter, the wavelet coefficients are amplified or attenuated using learned parameters in the wavelet domain whose amount is linearly proportional to the number of the input graph nodes. Thus, the learned graph wavelet filters are restricted to underlying graphs and their applicability to tasks with large-scale or varying-size graphs is limited.

In this paper, we propose to implement graph filters with adaptive graph wavelets to improve the model capacity and scalability and develop a novel class of graph neural networks. Inspired by the ability of lifting structures in adapting wavelets on graphs~\cite{narang2009lifting,NIPS2013_5046,hidane2013lifting}, we propose to learn adaptive graph wavelet transforms with neural network-parameterized lifting structures that mainly consist of two elementary operations, i.e., prediction and update operations. The two operations are implemented with structure-aware attention to jointly consider the graph topology and node features. To cope with the loss of the edges (i.e., \emph{structural information loss}) in partitioning non-bipartite graphs during lifting~\cite{narang2009lifting}, we propose to lift based on diffusion wavelets that can well preserve the local structural information~\cite{donnat2018learning}. The sparse and local diffusion wavelets together with the proposed structure-aware lifting operations that propagate information between directly-connected nodes guarantee the locality and sparsity of the learned graph wavelets. The scalability of the lifting structures is achieved by learning  lifting operations with attention such that the number of learnable parameters is independent of the graph size. To improve the scalability and interpretability of graph wavelet filters, we further derive a soft-thresholding filtering operation by learning sparse graph representations in terms of the wavelets, avoiding the parameter-intensive and graph-dependent graph wavelet filters~\cite{xu2018graph,li2020fast,zheng2021how}. Moreover, to ensure that the learned graph representations are invariant to node permutations (i.e., permutation invariance), a permutation-invariant layer is further employed at the input of the network to sort the graph nodes into a permutation-invariant order according to a metric that can be efficiently calculated with diffusion wavelets. This metric captures the local topology information around each node, which is invariant to node permutations. We evaluate the proposed networks in node-level and graph-level representation learning tasks on benchmark graph datasets including citation networks (i.e., \emph{Cora}, \emph{Citeseer} and \emph{Pubmed}) and bioinformatics datasets (i.e., \emph{PROTEINS}, \emph{DD}, \emph{NCI1}, \emph{NCI109}, and \emph{Mutagenicity}). Extensive experimental results demonstrate the superiority of the proposed lifting-based adaptive graph wavelet networks over existing SGNNs. To sum up, our contributions are three folds:
\begin{itemize}
\item We propose a novel class of graph neural networks that implements localized, efficient, and scalable graph filters with adaptive graph wavelets and soft-thresholding filtering.
\item We propose novel structure-aware attention-based lifting operations to jointly exploit the graph structures and node features in constructing adaptive graph wavelets while guaranteeing the locality and sparsity of the resulting wavelets as well as the scalability of the graph wavelet filters for large and varying-size graphs. 
\item Lifting-based adaptive graph wavelet networks are developed to learn permutation-invariant representations for node and graph classification, demonstrating superior or comparable performance with the state-of-the-arts.
\end{itemize}

The rest of this paper is organized as follows. We present related work about graph neural networks in Section~\ref{sec:related}. In Section~\ref{sec:pre}, we provide some preliminaries on graph Fourier and wavelet transforms, Fourier-based and wavelet-based graph filters, and lifting structures. In Section~\ref{sec:gconv}, we present the proposed lifting-based adaptive graph wavelet filters and its properties (i.e., locality, sparsity, and vanishing moments). In Section~\ref{sec:GFCN}, lifting-based adaptive graph wavelet networks are developed for permutation-invariant graph representation learning. In Section~\ref{sec:exp}, we evaluate the proposed models with extensive experiments on benchmark citation and bioinformatics graph datasets for node and graph classification tasks. Finally, We conclude this paper in Section~\ref{sec:con}.
 
\section{Related Work}\label{sec:related}
The striking success achieved by convolutional neural networks on Euclidean domains has ignited the enormous interest in developing graph neural networks~\cite{bronstein2017geometric,zhou2020graph,wu2020comprehensive,zhang2020deep} for non-Euclidean data that can be represented with graphs. Existing graph neural networks can be roughly classified into two categories, i.e., spatial-based and spectral-based models. Spatial-based graph neural networks generally aggregate information from neighborhood of graph nodes. Early works (e.g., GraphSAGE~\cite{hamilton2017inductive}, DiffGCN~\cite{atwood2016diffusion}) sample a fixed number of neighboring nodes according to some distance metrics (e.g., shortest-path distance or diffusion distance) to facilitate the learning of weight-sharing spatial filters. Recently, most of the spatial models follow the message passing scheme~\cite{gilmer2017neural}. For example, GAT~\cite{velivckovic2017graph} adopts a self-attention mechanism to learn
anisotropic weights for message passing. To incorporate high-order graph information, advanced feature extraction schemes e.g., JKnet~\cite{xu2018representation}, MixHop~\cite{abu2019mixhop}, APPNP~\cite{klicpera2018predict} and GDC~\cite{klicpera2019diffusion} are further proposed to leverage dense/residual connections or diffusion/random walk in order to facilitate multi-scale and long-range message passing. To improve the generalization performance and robustness of the GNNs, some advanced topological denoising~\cite{luo2021learning}, graph sparsification~\cite{zheng2020robust}, and data augmentation approaches~\cite{zhao2021data} are further leveraged to drop potentially tasks-irrelevant edges from the input graphs to improve graph representation learning.

In contrary to spatial-based models, spectral-based graph neural networks are grounded on spectral graph theory~\cite{chung1997spectral} and graph signal processing~\cite{ortega2018graph}. The pioneering works~\cite{shuman2013emerging,bruna2013spectral} define graph filters in the graph Fourier domain spanned by the eigenvectors of the graph Laplacian operator. However, the graph Fourier bases are dense, global-supported, and fixed for underlying graphs, leading to computation, memory-intensive, and non-local graph filters. For efficient and localized graph filters, the spaces for the learnable graph filters are limited. For example, smooth graph filters are generally learned for better spatial locality according to the duality of the spectral and spatial domain~\cite{DBLP:journals/corr/HenaffBL15}. ChebNet~\cite{defferrard2016convolutional} achieves efficient computation and exact locality by learning graph filters represented with the low-order Chebyshev polynomials, which is further simplified by GCN~\cite{kipf2016semi} with 1-order polynomials for semi-supervised node classification. Recently, GCNII and its variants~\cite{chen2020simple } relieve the over-smoothing problem with initial residual connections and residual identity mapping and extend GCNs to truly deep nonlinear models. For more flexible spectral filter shapes (e.g., localized on narrow frequency bands), filters based on rational polynomials (e.g., Cayley polynomials~\cite{levie2018cayleynets}), Auto-regressive Moving Average filters~\cite{bianchi2021graph}, and feedback looped filters~\cite{asiri2019dfnets} are further developed. Another line of research proposes to implement graph filters with graph wavelet transforms to leverage their inherent sparsity and spectral and spatial locality. GWNN~\cite{xu2018graph} implements graph filters with diffusion wavelets~\cite{coifman2006diffusion}, where the wavelet coefficients are processed with parameter-intensive diagonal filters whose size depends linearly on the number of nodes of the underlying graphs. HANet~\cite{li2020fast} learns graph filters with Haar basis constructed based on a given chain of multi-resolution graphs~\cite{belkin2006manifold,chui2015representation}. To reduce the number of filter parameters, weight-sharing scheme is proposed to leverage the clustering structures within the given chain of graphs. Nevertheless, the design of a proper chain of coarsen-grained graphs is also a challenging problem in graph learning community. The most recent work, UFG~\cite{zheng2021how},  constructs graph filters with undecimated framelet transforms to leverage the low-pass and high-pass components and learn multi-scale features. However, the multi-scale and multi-band framelet transforms introduce significantly more computation and memory overheads in comparison to previous wavelet-based models. Moreover, the learnable parameters of the graph filters in the wavelet domain are still graph-dependent and intensive with a complexity of $\mathcal{O}(JnN)$, where $J$, $n$, $N$ are the number of scales, bands, and graph nodes, respectively. Notably, the work in~\cite{zheng2021how} also notices the benefits of soft-thresholding operations (i.e., wavelet-shrinkage), but utilizes it as an activation function following the learnable parameters to improve the representation robustness. In contrast, with adaptive graph wavelets, we adopt the soft-thresholding operation to perform filtering directly, avoiding the parameter-intensive and graph-dependent graph filters. Besides, there are some works that generalize scattering transforms~\cite{bruna2013invariant} to graphs for robust representations~\cite{zou2020graph,gama2018diffusion,gama2019stability} or incorporate band-pass information for more discriminative graph representation~\cite{min2020scattering}.

\section{Preliminaries}\label{sec:pre}
In this section, we briefly review graph Fourier and wavelet transforms, Fourier-based and wavelet-based graph filters, and lifting structures. 

\subsection{Graph Fourier and Wavelet Transforms}
Existing spectral graph convolutions are generally defined in transform domains (e.g., graph Fourier or graph wavelet domain). Denote ${\mathcal G}=({\mathcal V},{\mathcal E},\mathbf{W})$ as an undirected graph with the node set $\mathcal V$ and edge set $\mathcal E$, and $\mathbf{W}\in {\mathbb R}^{N\times N}$ is the adjacency matrix which encodes the pairwise relationships among the $N$ nodes. The normalized graph Laplacian matrix is calculated as $\mathbf{L}=\mathbf{I}-\mathbf{D}^{-1/2}\mathbf{W}\mathbf{D}^{-1/2}$, where $\mathbf{D}$ is the diagonal degree matrix with $\mathbf{D}_{ii}=\sum_j{\mathbf{W}_{ij}}$ and $\mathbf{I}$ is a identity matrix. For undirected graphs, $\mathbf{L}$ is a positive semi-definite matrix that has a complete set of orthogonal eigenvectors $\{\mathbf{u}_l\}_{l\in\{1,2,\cdots,N\}}$ with the corresponding ordered non-negative eigenvalues $\{\lambda_1,\lambda_2,\cdots,\lambda_N\}$. It can be rewritten as 
\begin{equation}\label{eq1}
\mathbf{L}=\mathbf{U}\mathbf{\Lambda}\mathbf{U}^T,
\end{equation}
where $\mathbf{U}=[\mathbf{u}_1,\mathbf{u}_2,\cdots, \mathbf{u}_N]$ is the Fourier basis and $\mathbf{\Lambda}=\mathrm{diag}(\lambda_1,\lambda_2,\cdots,\lambda_N)$ is the spectrum. For signal $\mathbf{x}$ defined on graph $\mathcal G$ with a scalar on each vertex, the graph Fourier transform is defined as $\mathbf{\hat{x}}=\mathbf{U}^T \mathbf{x}$ and the inverse Fourier transform is $\mathbf{x}=\mathbf{U}\mathbf{\hat{x}}$. Consequently, the spectral graph wavelets~\cite{hammond2011wavelets} can be defined in graph Fourier domain as
\begin{equation}\label{eq2}
\boldsymbol{\Psi}_t=\mathbf{U}\mathbf{G}_t\mathbf{U}^T,
\end{equation}
where $\mathbf{G}_t=\text{diag}(g(t\lambda_l))_{l\in {1,2,\cdots,N}}$ and $g(t\lambda)$ is some spectral graph wavelet generating kernel with scaling parameter $t$ (e.g., diffusion kernels or functions satisfying the wavelets admission condition~\cite{hammond2011wavelets}). For example, with the diffusion kernel $g(\lambda)=e^{-\lambda}$, the diffusion wavelets with scaling parameter $t$ can be constructed as
\begin{equation}\label{eq11}
\mathbf{\Psi}_t=\mathbf{U}e^{-t\mathbf{\Lambda}}\mathbf{U}^T.
\end{equation}
The wavelet transform and inverse wavelet transform could be implemented as $\mathbf{\hat{f}}=\mathbf{\Psi}_t^T\mathbf{f}$ and  $\mathbf{f}=\mathbf{\tilde{\Psi}}_t\mathbf{\hat{f}}$, respectively, where $\mathbf{\tilde{\Psi}_t}$ is the corresponding dual wavelets.

\subsection{Fourier-based and Wavelet-based Graph Filters}
Graph filters are originally defined in graph Fourier domain~\cite{shuman2013emerging} by generalizing the concepts in classical signal processing to graphs. Given a graph signal ${\mathbf x}\in {\mathbb{R}^N}$ defined on a graph $\mathcal{G}$, graph filtering with a graph filter $\mathbf{g} \in{\mathbb{R}^N}$ is 
\begin{equation}\label{eq3}
\mathbf{x}\star_{\mathcal{G}}\mathbf{g}=\mathbf{U}(\mathbf{U}^T\mathbf{g}\odot\mathbf{U}^T\mathbf{x})={\mathbf U}{\mathbf{\hat{g}}}{\mathbf U}^T{\mathbf x},
\end{equation}
where $\mathbf{\hat{g}}=\text{diag}({\mathbf U}^T{\mathbf g})$ is the corresponding filter in the spectral domain. Most of the existing graph filters follow this definition with different designs of  $\mathbf{\hat{g}}$. To leverage the locality and sparsity of graph wavelet transforms, graph wavelet filters~\cite{xu2018graph} are further developed to achieve graph filtering.
\begin{equation}\label{eq4}
\mathbf{x'}=\mathbf{\tilde{\Psi}_t} \mathbf{\hat{g}} \mathbf{\Psi}_t^T\mathbf{x}
\end{equation}

Existing graph wavelet networks~\cite{xu2018graph,li2020fast,zheng2021how} leverage predefined wavelets (e.g., diffusion, Haar or framelet wavelets) to implement graph wavelet filters and cannot adapt to signals and tasks at hand. Furthermore, graph filtering in Eq.~\eqref{eq4} is performed in the graph wavelet domains with a learnable multiplication operator $\hat{\mathbf{g}}$, where the number of learnable parameters is intensive and graph-dependent. Thus, these networks cannot scale to large-scale and varying-size graphs. In this paper, we propose to improve previous graph wavelet networks~\cite{xu2018graph,li2020fast,zheng2021how} with adaptive graph wavelets learned with neural network-parameterized lifting structures.

\subsection{Lifting Structures}
\begin{figure}[!t]
\renewcommand{\baselinestretch}{1.0}
\centering
\centering
\includegraphics[width=0.8\columnwidth,height=0.3\columnwidth]{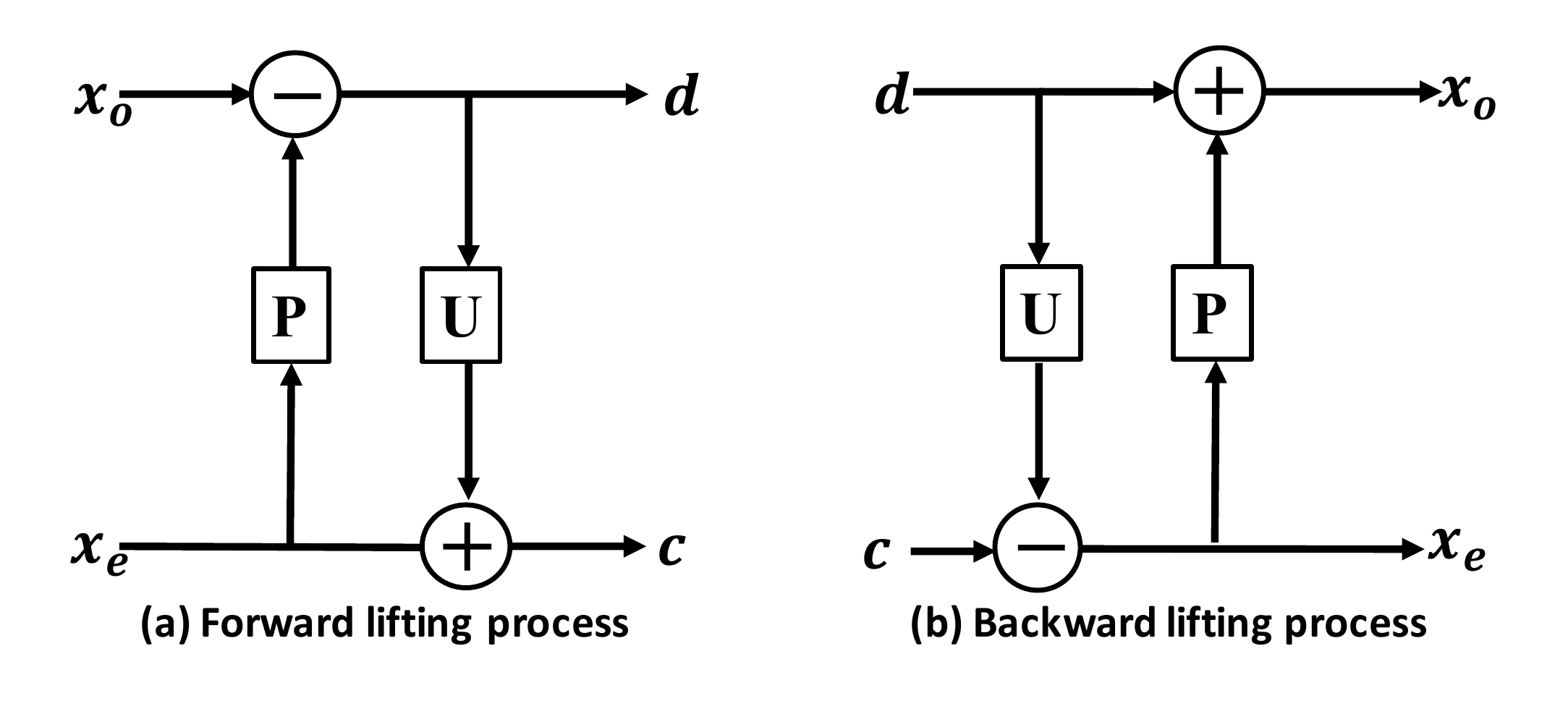}
\caption{Illustrative diagram of lifting scheme. (a) illustrates the forward lifting process with one lifting step (i.e., one prediction and one update step) and (b) presents the corresponding backward lifting process.}\label{fig1}
\end{figure}

\begin{figure}[!t]
\renewcommand{\baselinestretch}{1.0}
\centering
\includegraphics[width=0.8\columnwidth]{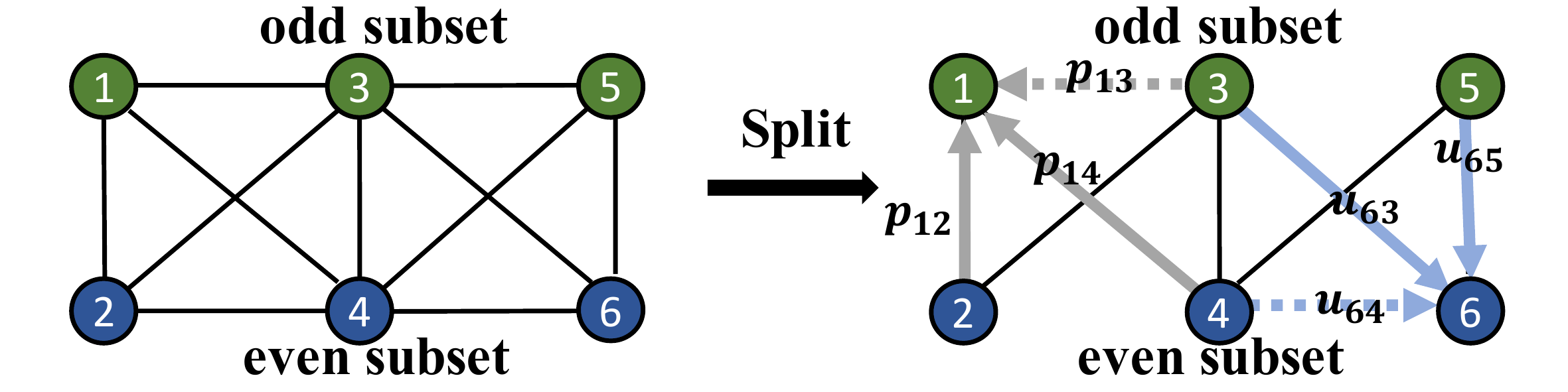}
\caption{Illustration of lifting on an example graph. Structural information is lost after splitting the graph (e.g., the edges between node 3 and 1, 6 and 4 are lost and not utilized to propagation information), which degrades the ability for capturing signal correlations.}\label{fig2}
\end{figure}

Classical wavelets defined on regular domains (e.g., 1-D, 2-D, or 3-D grids) are constructed by shifting and scaling mother wavelets. This construction cannot be adapted to irregular graphs in a straightforward way due to the lack of the intrinsic notions of translation and dilation. The lifting structure~\cite{SWELDENS1996186,sweldens1998lifting}, which factors the wavelet transforms into elementary steps, namely lifting steps, is a general framework to customize wavelets adapted to arbitrary domains (e.g., graphs, manifolds) or data/signals at hand, leading to the so-called second-generation wavelets. Its spatial implementation also leads to a computation- and memory-friendly calculation by performing wavelet transforms without explicitly calculating the wavelet basis. Lifting structures have been widely employed in constructing wavelets on graphs (or trees) for efficient data manipulations in sensor networks~\cite{shen2008optimized}, sparse representation~\cite{NIPS2013_5046} and denoising~\cite{narang2009lifting} for graph signals. Typically, a single lifting step consists of three stages: splitting, prediction, and update, as presented in \figurename~\ref{fig1}(a). To better illustrate the process of lifting in constructing wavelets on graphs, we take a simple graph (\figurename~\ref{fig2}) for example.
\begin{itemize}
\item {\bf Splitting:} The graph is divided into even and odd subsets, namely even subset $\mathbf{x}_e$ and odd subset $\mathbf{x}_o$. For simplicity, hereafter, we call the signals residing on the even and odd subsets even and odd coefficients, respectively. 
\item{\bf Prediction:} This stage is equivalent to performing a high-pass filtering and subsampling $\mathbf{x}$ for wavelet coefficients on the odd subset.
The wavelet coefficients are obtained by calculating the prediction residual between odd coefficients and its predictor
\begin{equation}\label{eq5}
\mathbf{d}[i]=\mathbf{x}_o[i]-\mathbf{P}(\mathbf{x}_e)[i],
\end{equation}
where $\mathbf{P}(\mathbf{x}_e)$ is the linear combination of $\mathbf{x}_e$ with the prediction weight $p_{ij}$ as $\mathbf{P}(\mathbf{x}_e)[i]=\sum_{i\sim j}p_{ij}{\mathbf x}_{e}[j]$.
\item{\bf Update:} A low-pass filtered and coarse version of $\mathbf{x}$ is obtained by updating the even coefficients $\mathbf{x}_e$ with $\mathbf{d}$.
\begin{equation}\label{eq6}
\mathbf{c}[j]=\mathbf{x}_e[j]+\mathbf{U}(\mathbf{d})[j],
\end{equation}
where $\mathbf{U}(\mathbf{d})$ is the linear combination of $\mathbf{d}$ as $\mathbf{U}(\mathbf{d})[j]=\sum_{j\sim i}u_{ji}{\mathbf d}[i]$ and $u_{ij}$ is the update weight. The approximation coefficients $\mathbf{c}$ correspond to the coefficients of a scaling transform.
\end{itemize}

The lifting process is invertible and the inverse wavelet transform cFan be performed as illustrated in \figurename~\ref{fig1}(b). Note that the above prediction and update step can be iterated down for more advanced wavelet transforms. Considering the transform stability of the resulting wavelet transform, in this paper, the commonly-used update-first scheme is employed~\cite{claypoole2003nonlinear, NIPS2013_5046}. In other words, we perform update before prediction. 

Though lifting structures are widely employed in constructing graph wavelets~\cite{shen2008optimized,NIPS2013_5046,narang2009lifting}, general graphs except the bipartite ones can not be trivially split into two disjoint subsets, leading to the so-called \emph{structural information loss} problem~\cite{narang2009lifting}, as shown in \figurename~\ref{fig2}. To maximally preserve the edge information (structural information) in lifting, the graph partition is commonly formulated as the well-known Maxcut problem, which is, however, NP-hard. Greedy or spectral clustering algorithms are typically employed. Nevertheless, those algorithms are time-consuming and the greedy algorithms tend to produce different results in various runs. Moreover, the lifting operations are either hand-designed~\cite{shen2008optimized,narang2009lifting} or parameter-intensive and graph-dependent. Hence, they can not scale to large and varying-size graphs~\cite{NIPS2013_5046}. In this paper, we propose an effective and scalable neural network-parameterized lifting structure to learn adaptive graph wavelets.

\begin{figure*}[!t]
\renewcommand{\baselinestretch}{1.0}
\centering
\includegraphics[width=2.0\columnwidth,height=0.25\columnwidth]{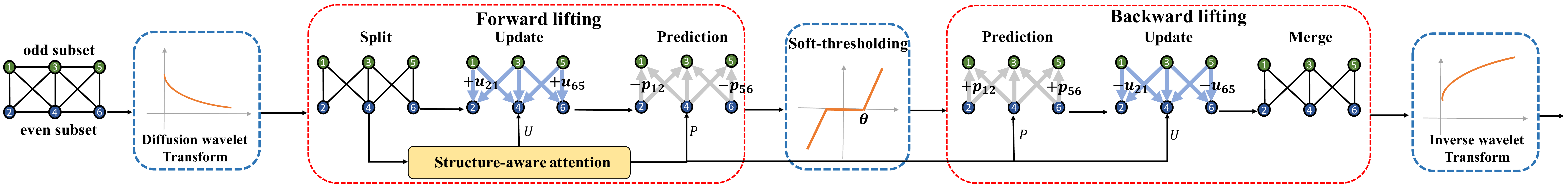}
\caption{The overview of the proposed lifting-based adaptive graph wavelet filters. The input graph signals are first transformed with diffusion wavelet transforms to smooth the node features according to the graph structural information. Then, the graphs are split into odd and even subsets. Next, structure-aware attention mechanism is adopted to estimate the update and prediction weights to construct new wavelets. The wavelet coefficients are then filtered with a soft-thresholding operation followed by the inverse lifting and wavelet transform for graph representations in the spatial domain.}\label{fig3}
\end{figure*}

\section{Lifting-based Adaptive Graph Wavelet Filters}\label{sec:gconv}
In this section, we propose lifting-based adaptive graph wavelet filters for graph wavelet networks. Denote $\mathbf{\Phi}$ as the basis of adaptive graph wavelet transforms (learned through training and are adaptive to signals and tasks at hand). Graph filtering with the adaptive graph wavelet transform is  
\begin{equation}\label{eq7}
\mathbf{x}'=\tilde{\mathbf{\Phi}}\mathbf{\hat{g}}\mathbf{\Phi}^T \mathbf{x},
\end{equation}
where $\mathbf{\Phi}^T\mathbf{x}$ is the learned forward wavelet transform, ${\tilde{\mathbf{\Phi}}}\mathbf{x}$ is the inverse wavelet transform, and $\mathbf{\hat{g}}$ is the corresponding filter in the wavelet domain. It is analogous to existing Fourier/wavelet-based graph filters except that it is constructed in the domain of learned graph wavelets $\mathbf{\Phi}$. 

\subsection{Design Choices}
We learn the adaptive graph wavelet transforms with neural network-parameterized lifting structures. To ensure that the resulting adaptive transforms learned with the lifting structure are localized, efficient, and satisfy the admissibility condition of spectral graph wavelets~\cite{hammond2011wavelets}, the learned transforms should have desirable properties like locality, sparsity, and with at least 1-order vanishing moment. Besides, the resulting graph wavelet filters need to scale to large-scale and varying-size graphs, i.e., scalability. 
In general, the challenges of constructing adaptive graph wavelet transforms with lifting structures are as follows:
\begin{itemize}
\item {\bf Structural information loss~\cite{narang2009lifting}.} The splitting of graphs (except bipartite graphs) would inevitably drop some edges, leading to the loss of structural information.
\item{\bf Locality and Sparsity.} The learned transforms should be sparse and spatially-localized such that the resulting graph filters are efficient and localized.
\item {\bf Vanishing moments.} The prediction and update operations should be well-designed such that the resulting wavelets are guaranteed to have at least 1-order vanishing moment, and consequently, satisfy the admissibility condition of spectral graph wavelets~\cite{hammond2011wavelets}.
\item {\bf Scalability.} The parameter complexity of lifting structures should be independent of the size of the input graphs such that they can scale to tasks with large and varying-size graphs.
\end{itemize}

We address the above challenges respectively with the following design choices.
\begin{itemize}
\item {\bf Lifting based on diffusion wavelets.} Instead of preserving more edges with non-deterministic and expensive greedy algorithms, we propose to perform diffusion graph wavelet transforms to encode the structural information around each node~\cite{donnat2018learning} before lifting. Diffusion graph wavelets are guaranteed to be spatially localized and sparse and can be efficiently calculated with Chebyshev polynomials~\cite{xu2018graph}. 
\item{\bf Structure-aware lifting operations} The lifting operations (i.e., prediction and update) are constrained to pass information between directly-connected nodes to guarantee the locality, which also ensures the sparsity of the resulting wavelet transforms for sparse graphs.
\item {\bf 1-order vanishing moment condition.} We propose a constraint (i.e., 1-order vanishing moment condition) on the update and predict operators (i.e.,$\mathbf{U}$ and $\mathbf{P}$) to guarantee that a single lifting step produces wavelets with $1$-order vanishing moment.
\item {\bf Attention-based lifting operations.} Instead of directly learning all prediction and update weights~\cite{NIPS2013_5046}, we apply attention mechanism to estimate the prediction and update weights such that graph structures and node features can be jointly considered and the number of parameters is independent of the graph size. 
\end{itemize}

As depicted in \figurename~\ref{fig3}, the proposed lifting-based adaptive graph wavelet filter consists of five components: diffusion wavelet transform, forward lifting process, wavelet filtering, backward lifting process, and inverse diffusion wavelet transform.

Graph filtering (i.e., Eq.~\eqref{eq7}) is achieved on multi-dimensional signals in a channel-wise fashion. For clarity, we describe the case of single-channel graph signals here and it can be further extended to the multi-channel case. We denote the input graph signal as $\mathbf{x}\in{\mathbb R}^{N}$, the diffusion wavelets and the dual ones with scaling parameter $t$ as $\mathbf{\Psi}_t$ and $\mathbf{\tilde{\Psi}}_t$ respectively, and the forward and backward lifting processes with $\mathbf{\Omega}$ and $\mathbf{\Omega}^{-1}$, respectively. The graph filtering with lifting-based adaptive graph wavelet filters can be formulated as
\begin{equation}\label{eq8}
\mathbf{x}'=\sigma(\mathbf{\tilde{\Psi}}_t\mathbf{\Omega}^{-1}(\mathbf{T}_\theta(\mathbf{\Omega}(\mathbf{\Psi}_t^T \mathbf{x})))),
\end{equation}
where $\sigma$ is a nonlinear activation function (e.g., ReLU or sigmoid) and $\mathbf{T}_\theta$ is a soft-thresholding filtering operation with hyper-parameter $\theta$ which will be elaborated in Section~\ref{St}. 

To elucidate this process, suppose that $\mathbf{\Omega}$ consists of a single lifting step. Given the coefficients $\mathbf{\hat {x}}$ after performing the diffusion wavelet transform  $\mathbf{\hat{x}}=\mathbf{\Psi}^T_t\mathbf{x}$, the forward lifting is then performed, where $\mathbf{\hat {x}}$ is randomly and equally split into $\mathbf{\hat{x}}_o$ and $\mathbf{\hat{x}}_e$ (in practice, we randomly split the underlying graphs by half and rearrange the graph nodes accordingly before feeding the graph data into models such that we can easily access odd and even coefficients as $\mathbf{\hat{x}}=[\mathbf{\hat{x}}_{o},\mathbf{\hat{x}}_{e}]$).
\begin{equation}\label{eq9}
\mathbf{\overline{x}}_e=\mathbf{\hat{x}}_e+\mathbf{U}(\mathbf{\hat{x}}_o),\quad\mathbf{\overline{x}}_o=\mathbf{\hat{x}}_o-\mathbf{P}(\mathbf{\overline{x}}_e),
\end{equation}
where $\mathbf{U}$ and $\mathbf{P}$ are update and predict operators, respectively.

Next, the resulting wavelet coefficients are filtered with a soft-thresholding operation as $\mathbf{\overline{x}}'_e=\mathbf{T}_{\theta}(\mathbf{\overline{x}}_e),\mathbf{\overline{x}}'_o=\mathbf{T}_{\theta}(\mathbf{\overline{x}}_o)$ followed by the inverse lifting process
\begin{equation}\label{eq10}
\mathbf{\hat{x}}'_o=\mathbf{\overline{x}}'_o+\mathbf{P}(\mathbf{\overline x}'_e),\quad\mathbf{\hat{x}}'_e=\mathbf{\overline{x}}'_e-\mathbf{U}(\mathbf{\hat{x}}'_o).
\end{equation}
Finally, $\mathbf{\hat{x}}'_o$ and $\mathbf{\hat{x}}'_e$ are merged as $\mathbf{\hat{x}}'=[\mathbf{\hat{x}}'_o,\mathbf{\hat{x}}'_e]$ and transformed back to the spatial domain, i.e., $\mathbf{{x}}'={\mathbf{\tilde{\Psi}}_t}\mathbf{\hat{x}}'$. In the following subsections, we will elaborate each design choice.

\subsection{Lifting Based on Diffusion Wavelets}
The lifting structures construct wavelets by exploiting the signal correlations between the odd (even) nodes and their even (odd) neighbors. Hence, the ability for lifting operations to accurately capture such correlations is crucial. For graph data, the structures are very important in modeling such correlations, i.e., the data on a node can be estimated from its close neighbors. However, the lifting process can only propagate information between the odd and even subsets and the edges within each subset will be inevitably dropped. This \emph{structural information loss} may significantly hamper the ability of lifting structures to capture signal correlations. We resort to diffusion wavelets to alleviate \emph{structural information loss}.

The diffusion wavelets with scaling parameter $t$ can be generated with heat kernel as $\mathbf{\Psi}_t=\mathbf{U}e^{-t\mathbf{\Lambda}}\mathbf{U}^T$. The scaling parameter $t$ controls the locality of the wavelet basis. From a spatial perspective, each diffusion wavelet is centered at a node and spreads unit energy over its neighbor. These diffusion patterns are able to characterize the local structural information around each node~\cite{donnat2018learning}. After diffusion wavelet transforms, each node aggregates the information from its neighboring nodes and the resulting node features are summaries of overlap local subgraphs. Hence, the correlations between these node features become stronger and easier to capture for lifting operations with incomplete neighboring nodes. From a spectral perspective, the diffusion wavelet transform performs as a low-pass filter that discards the high frequency components of input graph signals. Hence, the signals on a node may be easily estimated with incomplete neighbors. 

However, the eigen-decomposition of the graph Laplacian is computationally expensive with the complexity of $\mathcal{O}(N^3)$ for graphs with $N$ nodes and is prohibitive for large-scale graphs. Fortunately, as in~\cite{hammond2011wavelets,xu2018graph}, efficient polynomial approximation methods (e.g., Chebyshev polynomial approximation) can be adopted to construct diffusion wavelets for large-scale graphs and the computational complexity can be reduced to $\mathcal{O}(K|\mathcal{E}|)$, where $K$ is the order of polynomials and $|\mathcal{E}|$ is the number of edges in the graph. Note that most real-world graphs are sparse~\cite{chung2010graph} with $|\mathcal{E}|\ll N^2$. Thus, the diffusion wavelets can be efficiently calculated with Chebyshev polynomial approximation for large-scale graphs. Refer to the supplemental material for details of the Chebyshev polynomial approximation for diffusion wavelets.

The problem of \emph{structural information loss} can be relieved by performing diffusion wavelet transforms before lifting. We further avoid partitioning the graphs with expensive greedy algorithms. Instead, we randomly split the graphs into two halves such that approximately half of the edges are preserved. Experimental results demonstrate the desirable performance of proposed method in practice. 

\subsection{Structure-aware Attention-based Lifting Operations}
To ensure the locality and sparsity of the resulting graph wavelets, the update and prediction operations should be localized and sparse. We thus constrain them by propagating information between directly-connected odd and even nodes (i.e., \emph{structure-aware}). Though graph structures are important information for capturing the signal correlations, they cannot comprehensively reflect them. Therefore, modeling these correlations with the given graph structures (possibly with edge weights) may not lead to desirable wavelets. Besides, constructing wavelets with  graph structures solely cannot produce wavelets adapted to the signals and tasks at hand. Alternatively, learning all the prediction or update weights (i.e., $p_{ij}$ and $u_{ij}$) directly seems promising but it would result in parameter-intensive and graph-dependent lifting operations which are expensive for large and varying-size graphs and hard to train~\cite{NIPS2013_5046}. We use attention mechanisms~\cite{velivckovic2017graph} whose parameter complexity is independent of the size of input graphs to capture relationships between different objects and implement the lifting structures efficiently. Most importantly, graph structures and signals are jointly considered.

Specifically, we randomly split a graph with $N$ nodes and rearrange the data $\mathbf{x}\in \mathbb{R}^{N\times d}$ on this graph as $\mathbf{x}=[\mathbf{x}_o,\mathbf{x}_e]$, where $\mathbf{x}_o$ and $\mathbf{x}_e$ denote the data on odd and even nodes, respectively. The adjacency matrix can be rearranged as
\begin{equation}\label{eq12}
\mathbf{W}=\begin{bmatrix}
\mathbf{O}_{\frac{N}{2}\times\frac{N}{2}} & \mathbf{K}_{\frac{N}{2}\times\frac{N}{2}} \\
\mathbf{Q}_{\frac{N}{2}\times\frac{N}{2}} & \mathbf{E}_{\frac{N}{2}\times\frac{N}{2}} \\
\end{bmatrix},
\end{equation}
where $\mathbf{O}$ and $\mathbf{E}$ are the adjacency matrices of the odd and even subgraphs while $\mathbf{K}$ and $\mathbf{Q}$ are the sub-matrices connecting nodes across the odd and even subsets which reflect their interactions. Note that $\mathbf{Q}=\mathbf{K}^T$ for undirected graphs. 

The relationships between pairwise odd and even nodes are learned with attention mechanism. We first sample the adjacency matrix by keeping the connections across the odd and even subsets as
\begin{equation}\label{eq13}
\mathbf{W_s}=\begin{bmatrix}
\mathbf{0}_{\frac{N}{2}\times\frac{N}{2}} & \mathbf{K}_{\frac{N}{2}\times\frac{N}{2}} \\
\mathbf{Q}_{\frac{N}{2}\times\frac{N}{2}} & \mathbf{0}_{\frac{N}{2}\times\frac{N}{2}} \\
\end{bmatrix}.
\end{equation}
Then, the structure-aware attention mechanism~\cite{velivckovic2017graph} is adopted to learn the prediction and update weights to model the correlations between odd and even nodes as
\begin{equation}\label{eq14}
\mathbf{W}_{a,ij}=\mathbf{a_1}[\mathbf{a_2x_i}\|\mathbf{a_2x_j}], \quad \text{if}\ \mathbf{W}_{s,ij}>0,
\end{equation}
where $\mathbf{x}_i\in\mathbb{R}^d$ is the feature of the $i$-th node, $\mathbf{a}_1\in \mathbb{R}^{2c}$ and $\mathbf{a}_2 \in \mathbb{R}^{c\times d}$ are the learnable parameters shared for all nodes, and $\|$ denotes concatenation along the channel dimension. This process produces a new adjacency matrix $\mathbf{W}_a$ that could better model the correlations between the two subsets in a data-driven manner. The  sub-matrices of $\mathbf{W}_a$ that model the correlations between odd and even subsets are then utilized to calculate predict and update operators and are denoted as $\mathbf{K}_a$ and $\mathbf{Q}_a$.

Finally, to ensure that a single lifting step formulated as Eq.~\eqref{eq9} is able to produce wavelets with at least $1$-order vanishing moment, the update and predict operators $\mathbf{U}$ and $\mathbf{P}$ are required to satisfy the \emph{1-order vanishing moment condition}.
\begin{equation}\label{eq15}
\quad \mathbf{U1}=\mathbf{1},\quad \mathbf{P1}=\frac{\mathbf{1}}{\mathbf{2}},
\end{equation}
where $\mathbf{1}$ and $\mathbf{\frac{1}{2}}$ are constant vectors. In practice, the update and predict operators $\mathbf{U}$ and $\mathbf{P}$ can be calculated and normalized with Softmax operation to guarantee that the \emph{1-order vanishing moment condition} is satisfied: 
\begin{equation}\label{eq16}
\mathbf{U}=\text{Softmax}(\mathbf{Q}_a),\quad \mathbf{P}=\frac{1}{2}\text{Softmax}(\mathbf{K}_a).
\end{equation}

The structure-aware attention-based lifting operations can guarantee the locality, sparsity, 1-order vanishing moment, and scalability of lifting-based wavelets, as proven in Section~\ref{properties}. 

\subsection{Soft-thresholding Filtering}\label{St}
In contrary to existing graph wavelet filters that learn less interpretable graph representations by selectively amplifying or attenuating wavelet coefficients with parameter-intensive and graph-dependent multiplication operators~\cite{xu2018graph,li2020fast,zheng2021how}, we propose to learn sparse graph representations in terms of the learned adaptive wavelets through soft-thresholding filtering operations, which is more interpretable, efficient, and can scale to tasks with large and varying-size graphs. 

This design is inspired by the most recent work~\cite{zhu21}  that interprets existing graph filters (e.g., \cite{kipf2016semi,wu2019simplifying,xu2018representation,klicpera2018predict}) with an optimization framework
\begin{equation}\label{eq18}
\mathbf{X'}=\text{argmin}_{\mathbf{X'}}(\|\mathbf{F}_1\mathbf{X'}-\mathbf{F}_2\mathbf{X}\|_F^2+\lambda\text{Tr}(\mathbf{X'}^T\mathbf{L}\mathbf{X'})),
\end{equation}
where $\lambda$ is a regularization coefficient controlling the influence of different terms, $\mathbf{F}_1$ and $\mathbf{F}_2$ are arbitrary graph convolutional kernels, and $\mathbf{X}$ is the input graph signal. In Eq.~\eqref{eq18}, the first term is a feature fitting term that encodes the information of input graph signals and the second term is the graph Laplacian regularization that captures the homophily by enforcing the smoothness of the learned representations in terms of the underlying graph. For example, GCN~\cite{kipf2016semi} can be reformulated according to Eq.~\eqref{eq18} by setting $\mathbf{F}_1=\mathbf{F}_2=\mathbf{I}$ and $\lambda=1$ as
\begin{align*}
\mathbf{X'}=\text{argmin}_{\mathbf{X'}}(\|\mathbf{X'}-\mathbf{X}\|_F^2+\text{Tr}(\mathbf{X'}^T\mathbf{L}\mathbf{X'})),
\end{align*}
which learns a smooth approximation of $\mathbf{X}$.
Different convolutions build the relationships between $\mathbf{X}$ and $\mathbf{X}'$ in different spaces (e.g., original, low-frequency and high-frequency) with different convolutional kernels, but they all learn global smooth graph representations due to the graph Laplacian regularization term and high-frequency components of the input graph signals are attenuated globally. However, the homophily of local subgraphs varies across the graph. For example, the nodes in the center of a cluster exhibit more homophily with its neighbors than those in edges of clusters. Therefore,  global smoothness regularization is not necessarily an optimal choice.

Graph wavelets are localized in both spatial and spectral domains, which could filter signals supported on different local subgraphs differently. With the transform $\mathbf{\Phi}$ (consists of scaling and wavelet transforms) learned with our lifting structures that is able to adaptively filter local graph signals in the spectral domain, we propose to replace the smoothness regularization in Eq.~\eqref{eq18} with sparsity regularization for sparse wavelet representations
\begin{equation}\label{eq19}
\mathbf{X'}=\text{argmin}_{\mathbf{X'}}(\|\mathbf{\Phi}^T\mathbf{X'}-\mathbf{\Phi}^T\mathbf{X}\|_F^2+\theta\|
\mathbf{\Phi}^T \mathbf{X'}\|_1)),
\end{equation}
where $\theta$ is a non-negative coefficient that controls the influence of the sparsity regularization term. Note that instead of encoding the information of the input graph signal $\mathbf{X}$ into the filtered graph representation $\mathbf{X}'$ in the signal domain as is GCN (i.e., $\mathbf{F}_1=\mathbf{F}_2=\mathbf{I}$), we encode these information in the wavelet domain, such that high-frequency regularity of $\mathbf{X}$ captured by the learned wavelets can be well preserved.
Eq.~\eqref{eq19} indicates that the learned representation is a sparse representation of the input graph signal $\mathbf{X}$. In such a way, smoothness (i.e., homophily) can be captured locally via the learnable graph wavelets. 

The close-form solution of Eq.~\eqref{eq19} leads to the proposed soft-thresholding wavelet filtering  
\begin{equation}\label{eq20}
\mathbf{X'}=\mathbf{\tilde{\Phi}}{\mathbf T}_{\theta}(\mathbf{\Phi}^T\mathbf{X}),
\end{equation}
where $\tilde{\mathbf{\Phi}}$ is the inverse wavelet transform and  $\mathbf{T}_{\theta}$ is 
\begin{align*}\label{eq21}
 \mathbf{T}_{\theta}(y)=
 \left\{
 \begin{array}{ll}
 \text{sign}(y)(|y|-\theta), &\quad |y|>\theta\\
 0, &\quad \text{otherwise},
 \end{array}
 \right.
\end{align*} 
where $\theta$ is a non-negative hyper-parameter. 

By replacing the parameter-intensive and graph-dependent multiplication operator with the soft-thresholding filtering operation, we can significantly improve the scalability and improve the interpretability via the sparse graph representations.

\subsection{Properties of the Lifting-based Wavelets}\label{properties}
We now show the locality and sparsity of the graph wavelet transforms learned with the proposed lifting structures and then prove that a single lifting step (i.e, Eq.~\eqref{eq9} and Eq.~\eqref{eq15}) produces wavelets with $1$-order vanishing moment.

\subsubsection{Locality and Sparsity} 
The locality of the resulting transforms is influenced by the locality of the diffusion graph wavelet transforms and the subsequent lifting process. Therefore, we first analyze the locality of diffusion wavelets and then study the locality of the proposed lifting structure. As $k$-order polynomial graph filters are exactly localized within $k$-hop neighboring nodes of the underlying graphs~\cite{hammond2011wavelets}, spectral graph wavelets are typically approximated with polynomial graph filters to study their localization. 

We start by defining the spectral graph wavelet localized on a single node. Note that the matrix consisting of spectral graph wavelets in a single scale (i.e., $\mathbf{\Psi}_t$ in~\eqref{eq2}) is symmetric. For simplicity, we omit the transpose in the definitions and theorems when applying wavelet transforms to graph signals. 
\begin{definition}
\emph{Denote $\mathbf{\Psi}_t$ as the spectral graph wavelet with scaling parameter $t$. The wavelet localized on the $m$-th node $\mathbf{\Psi}_{t,m}$ is defined by applying the spectral graph wavelet transform to impulse on the $m$-th node i.e., $\mathbf{\Psi}_{t,m}=\mathbf{\Psi}_t\mathbf{\delta}_m$.}
\end{definition}

Lemma 1 shows that, if the error between the spectral graph wavelet kernel and its $k$-order polynomial approximation is bounded by a function of the scaling parameter $t$ on the graph spectrum, the error between the spectral graph wavelet $\mathbf{\Psi}_{t,m}$ and its $k$-order polynomial approximation is bounded by the same function.

\begin{lemma} \label{lemma1}
\emph{Let $\mathbf{L}$ be a normalized Laplacian matrix of graph $\mathcal{G}$. Denote $g(\lambda)$ as the wavelet kernel defined on graph spectrum $[0,\lambda_N]$, $p_k$ a $k$-order polynomial kernel. We further define $h(\lambda)$ the error between $g(\lambda)$ and $p_k$ i.e.,  $h(\lambda)=|g(\lambda)-p_k(\lambda)|$. Given any scale $t$, if ${\forall \lambda\in [0,\lambda_N]}, h(t\lambda)\leq \xi(t)$, we have that $\|h(t\mathbf{L})\|_2=\|g(t\mathbf{L})-p_k(t\mathbf{L})\|_2\leq \xi(t)$. Moreover,  $\|h(t\mathbf{L})\mathbf{\delta}_m\|_2=\|g(t\mathbf{L})\mathbf{\delta}_m-p_k(t\mathbf{L})\mathbf{\delta}_m\|_2\leq \xi(t)$, where $\delta_m$ is the dirac function at any node $m$.}
\begin{IEEEproof}
Please refer to supplemental material.
\end{IEEEproof}
\end{lemma}

In Theorem~\ref{thm2}, we present the approximation error between a diffusion wavelet localized on a single node and its $K$-order polynomial approximation.
\begin{thm}\label{thm2}
\emph{Let $g(t\lambda)$ be the diffusion wavelet kernel defined on graph spectrum $[0,\lambda_N]$ that generates $\mathbf{\Psi}_t$, $p_K(t\lambda)$ be the $K$-order Taylor polynomial approximation of $g(t\lambda)$ that generates the approximation wavelets $\tilde{\mathbf{\Psi}}_t$. Then, for arbitrary ${\lambda\in [0,\lambda_N]}$, we have $|g(t\lambda)-p_K(t\lambda)|\leq\frac{\lambda_N^{K+1}}{(K+1)!}t^{K+1}$.  Moreover, for diffusion wavelet localized on $m$-th node $\mathbf{\Psi}_{t,m}$ and its $K$-order approximation $\tilde{\mathbf{\Psi}}_{t,m}$, we have
\begin{equation}\label{eq17}
\frac{\|\mathbf{E}_{t,m}\|_2}{\|\mathbf{\Psi}_{t,m}\|_2}\leq \frac{\frac{\lambda_{N}^{K+1}}{(K+1)!}t^{K+1}}{1+\sum_{k=1}^K\frac{(-t)^k}{k!}\|L^k\delta_m\|_2-\frac{\lambda_{N}^{K+1}}{(K+1)!}t^{K+1}},
\end{equation}
where $\mathbf{E}_{t,m}=\mathbf{\Psi}_{t,m}-\mathbf{\tilde{\Psi}}_{t,m}$ is the approximation error of the diffusion wavelet and its $K$-order polynomial approximation.}
\begin{IEEEproof}
Please refer to supplemental material.
\end{IEEEproof}
\end{thm}

Theorem~\ref{thm2} shows that when $t \rightarrow 0$,  $\frac{\|\mathbf{\Psi}_{t,m}-\mathbf{\tilde{\Psi}}_{t,m}\|_2}{\|\mathbf{\Psi}_{t,m}\|_2}\rightarrow 0$. Therefore, when $t$ gets smaller, the wavelet can be more accurately approximated by its $K$-order polynomial approximation which is precisely localized within $K$-hops of the underlying graphs. For sparse graphs (i.e., graphs with sparse adjacency matrix), the locality of diffusion graph wavelets also implies their sparsity (real-world graphs are typically sparse~\cite{chung2010graph}).

Subsequently, we study the locality and sparsity of transforms realized with the proposed lifting structures. Since each of the proposed structure-aware attention-based operations (i.e., prediction and update operations) defined in Eq.~\eqref{eq9} and Eq.~\eqref{eq15} propagates information between directly-connected nodes, the prediction and update operations are both localized within one hop of neighboring nodes on the underlying graphs. Consequently, a single lifting step consists of the proposed structure-aware attention-based lifting operations is localized within two hops. Regarding sparsity, as the transforms realized with the proposed lifting structures are localized on the graphs, for an input graph with a sparse adjacency matrix and moderate expansion ratio, the resulting wavelets are sparse.

\subsubsection{Vanishing Moments}
Theorem~\ref{thm4} states that a single lifting step with the proposed lifting operations produces wavelets with 1-order vanishing moment~\cite{NIPS2013_5046}.
\begin{thm}\label{thm4}
\emph{For any constant graph signals $\mathbf{c}=c\mathbf{1}\in\mathbb{R}^N$, where $\mathbf{1}$ is the vector whose entries are all $1$, the wavelet coefficients (transform coefficients on the odd subset) produced by the wavelet transforms via a single lifting step  (Eq.~\eqref{eq9} and Eq.~\eqref{eq15}) are all zeros. That is the wavelets resulting from a single lifting step have 1-order vanishing moment.}
\begin{IEEEproof}
 The update and predict operators are constrained to satisfy Eq.~\eqref{eq15}. For any constant graph signal $\mathbf{c}=c\mathbf{1}\in\mathbb{R}^N$ that are spilt into $\mathbf{c_e}$ and $\mathbf{c}_o$, according to Eq.~\eqref{eq9} and Eq.~\eqref{eq15}, we have $\overline{\mathbf{c}}_e=\mathbf{c}_e+\mathbf{U}\mathbf{c}_o=2\mathbf{c}_e$ and $\overline{\mathbf{c}}_o=\mathbf{c}_o-\mathbf{P}\overline{\mathbf{c}}_e=\mathbf{0}_o$, where $\mathbf{0}_o$ denotes zero vector supported on the odd subset. So the wavelet coefficients on the odd subset are all zeros.
\end{IEEEproof}
\end{thm}

\section{Adaptive Graph Wavelet  Networks}\label{sec:GFCN}
To learn deep representations for multi-dimensional graph data, we now develop graph wavelet networks by stacking the proposed lifting-based graph wavelet filtering layers. However, there are two limitations that need be further addressed. First, the representations learned with the proposed graph filters are not invariant to the node permutations. Second, the proposed lifting-based graph wavelet filters process the input graph signal channel-wise. We address these issues in this section.

\subsection{Permutation Invariant Layer}\label{sub:pi}
We develop a permutation-invariant layer at the input of the network to rearrange the graph nodes in a canonical order. Diffusion wavelets (which have been precomputed for our lifting-based graph wavelet filters) reveal how a node see its local structure in an egocentric view. Therefore, they are able to characterize the information of local structure of each node~\cite{donnat2018learning}, which is invariant to node permutations. In this paper, we adopt the smoothness of the diffusion wavelets to sort the input graph nodes. 
\begin{definition}[Smoothness of diffusion wavelets]
\label{definition3}
\emph{Recall that $\mathbf{L}$ is the graph Laplacian. Let $\mathbf{\Psi}$ denote diffusion graph wavelets and $\mathbf{\Psi}_i$ represent the diffusion wavelet localized on node $i$. The smoothness of $\mathbf{\Psi}_i$ is defined as $s_i={\mathbf{\Psi}^T_i}\mathbf{L}\mathbf{\Psi}_i$.}
\end{definition}

The smoothness of all the graph wavelets localized on different nodes $\mathbf{s}=[s_1,s_2,\cdots,s_N]^T$ can be easily calculated by gathering the diagonal elements of $\mathbf{S}={\mathbf{\Psi}^T}\mathbf{L}\mathbf{\Psi}.$  We then prove in Theorem~\ref{thm5} that these smoothness values are permutation-equivariant.
\begin{thm}\label{thm5}
\emph{Let $g(\lambda)$ be the kernel generating the diffusion wavelets $\mathbf{\Psi}$ (i.e., $\mathbf{\Psi}=g(\mathbf{L})$), and $\mathbf{\Pi}$ be a permutation matrix. The graph Laplacian and diffusion wavelets of the permuted graph (i.e., $\mathbf{L}_{\mathbf{\Pi}}$ and $\mathbf{\Psi}_{\mathbf{\Pi}}$) can be calculated as $\mathbf{L}_{\mathbf{\Pi}}=\mathbf{\Pi}\mathbf{L}\mathbf{\Pi}^T$ and $\mathbf{\Psi}_{\mathbf{\Pi}}=\mathbf{\Pi} \mathbf{\Psi}\mathbf{\Pi}^T$, respectively. Then, we have $\mathbf{s}_{\mathbf{\Pi}}=\mathbf{\Pi}\mathbf{s}$, where $\mathbf{s}$ and $\mathbf{s}_{\mathbf{\Pi}}$ are the smoothness values of diffusion wavelets on the original graph and the permuted one, respectively.}
\begin{IEEEproof}
According to the definition of node permutations, we can directly formulate that $\mathbf{L}_{\mathbf{\Pi}}=\mathbf{\Pi}\mathbf{L}\mathbf{\Pi}^T$. Since the spectrum of $\mathbf{L}$ is discrete with $N$ values, the kernel $g(\lambda)$ defined on the graph spectrum can be exactly represented by a $N$-order polynomial function via interpolation. Therefore, the diffusion wavelets can be exactly represented with a polynomial graph filters $\mathbf{\Psi}=g(\mathbf{L})=p_N(\mathbf{L})$ and $\mathbf{\Psi}_{\mathbf{\Pi}}=g(\mathbf{L}_{\mathbf{\Pi}})=p_N(\mathbf{L}_{\mathbf{\Pi}})$. Since $\mathbf{L}_{\mathbf{\Pi}}=\mathbf{\Pi}\mathbf{L}\mathbf{\Pi}^T$ and $\mathbf{\Pi}^T\mathbf{\Pi}=\mathbf{I}$, so $p_N(\mathbf{L}_\mathbf{\Pi})=\mathbf{\Pi}p_N(\mathbf{L})\mathbf{\Pi}$, and then we have $\mathbf{\Psi}_{\mathbf{\Pi}}=\mathbf{\Pi} \mathbf{\Psi}\mathbf{\Pi}^T$. For the smoothness $\mathbf{s}_{\mathbf{\Pi}}$, we have 
\begin{align}\label{eq23}
\mathbf{s}_{\mathbf{\Pi}}&=\text{diag}(\mathbf{S}_{\mathbf{\Pi}})={\mathbf{\Psi}^T_{\mathbf{\Pi}}}\mathbf{L_{\mathbf{\Pi}}}\mathbf{\Psi}_{\mathbf{\Pi}}\nonumber\\
&=\text{diag}(\mathbf{\Pi} \mathbf{\Psi}^T \mathbf{\Pi}^T\mathbf{\Pi} \mathbf{L} \mathbf{\Pi}^T \mathbf{\Pi} \boldsymbol{\Psi} \mathbf{\Pi}^T)\nonumber\\
&=\text{diag}(\mathbf{\Pi}\mathbf{\Psi}^T \mathbf{L}\mathbf{\Psi} \mathbf{\Pi}^T)=\text{diag}(\mathbf{\Pi} \mathbf{S}\mathbf{\Pi}^T)\nonumber\\
&=\mathbf{\Pi}\text{diag}(\mathbf{S})=\mathbf{\Pi}\mathbf{s}.
\end{align}
The smoothness is permutation-equivariant.
\end{IEEEproof}
\end{thm}

Since the smoothness of diffusion wavelets is permutation-equivariant, we then arrange the nodes (in terms of the order of nodes in the input graph data vector $\mathbf{x}\in \mathbb{R}^{N\times d}$) according to the ascending order of these smoothness values. Therefore, the resulting graph node order in the input graph data is invariant to node permutations in graphs. When graph signals associated with the reordered graph nodes are fed into the cascaded lifting-based graph wavelet filtering layers, we can guarantee the permutation invariance of splitting of graphs and learned representations. Note that, though it requires $\mathcal{O}(N^2)$ computational complexity to calculate the smoothness of diffusion wavelets on graphs with $N$ nodes, the smoothness values are computed only once for each graph. In practice, we can precompute the smoothness and reorder the graph nodes as a data preprocessing step to improve training efficiency.

\subsection{Feature Transformation Layer}\label{sub:ft}
In the proposed graph wavelet filters, graph filtering is performed on each feature channel separately. The interactions among feature channels in classical convolutional neural networks are exploited by learning convolutional kernels for each pair of input and output feature channel, leading to a considerately large number of parameters and heavy computational and memory overheads. To alleviate this problem, we adopt the detached strategy as in~\cite{xu2018graph}. Features are first transformed with a feature transformation layer to explore interactions across channels before filtering.
Specifically, the multi-channel graph signal $\mathbf{X}^{l}\in \mathbb{R}^{N\times d_1}$ is transformed into $\mathbf{\hat{X}}^{l}\in \mathbb{R}^{N\times d_2}$ before being fed into lifting-based graph filters as
\begin{equation}\label{eq24}
\mathbf{\hat{X}}^{l}=\mathbf{X}^l\mathbf{W}
\end{equation}
where $\mathbf{W}\in {\mathbb R}^{d_1\times d_2}$ is a learnable parameter matrix.
Through this strategy, the number of the parameter as well as the computation complexity are significantly reduced.

\begin{figure}[!t]
\renewcommand{\baselinestretch}{1.0}
\centering
\includegraphics[width=0.9\columnwidth,height=0.3\columnwidth]{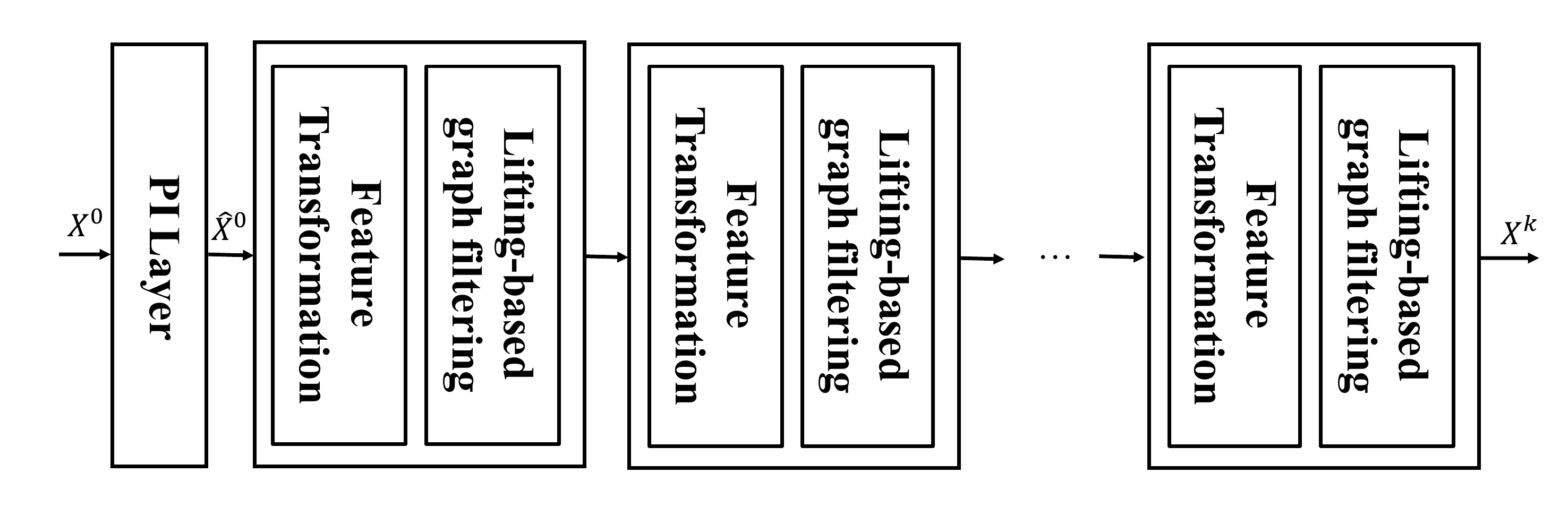}
\caption{The overview of the proposed adaptive graph wavelet network.}\label{fig4}
\end{figure}

Consequently, the proposed lifting-based graph wavelet network (LGWNN) consists of a permutation invariant layer at the input of the networks and stacked lifting-based graph wavelet filtering layers, each of which is composed of a feature transformation layer and a lifting-based graph wavelet filtering layer, as depicted in \figurename~\ref{fig4}.

\subsection{Model Complexity Analysis}
Here, we analyze the computation and parameter complexity of the proposed lifting-based adaptive graph wavelet filters. The calculation of diffusion wavelets and the smoothness values for graph reordering are performed once. Therefore, we precompute the diffusion wavelets and their smoothness values and reorder the graph accordingly to construct an augmented graph datasets before training the models. Specifically, for small graphs, eigen-decomposition is performed to calculate the diffusion wavelets. For large-scale graphs, efficient Chebyshev polynominal approximations can be adopted instead to generate the diffusion wavelets~\cite{hammond2011wavelets,defferrard2016convolutional,xu2018graph}. The computational complexity can be reduced to $\mathcal{O}(K|\mathcal{E}|)$, for $K$-order of polynomials and the graph with $|\mathcal{E}|$ edges. 

For each graph filtering layer, let $\mathbf{x}\in {\mathbb R}^{N\times d_{in}}$ denote the input graph signal with N nodes and $d_{in}$ feature channels, and $\mathbf{y}\in {\mathbb R}^{N\times d_{out}}$ denotes the output graph feature, the computation complexity is ${\mathcal O}(Nd_{in}d_{out}+ kN^2d_{out}+m|{\mathcal E}_{eo}|d_{out})$, where $|\mathcal{E}_{eo}|$ is the number of edges across the odd and even subsets (for random splitting, $|{\mathcal E}_{eo}|\approx \frac{1}{2}|{\mathcal E}|$), $m$ is the number of lifting steps (we adopt 1 lifting step i.e., $m=1$), and $k$ is the sparsity of the adopted diffusion wavelet basis with $k\ll 1$. The parameter complexity of each lifting-based graph filter is only ${\mathcal O}(d_{in} d_{out}+d_{out})$, independent of the graph size. Therefore, it can scale to tasks with large and varying-size graphs. Compared to previous graph wavelet filters, the proposed graph wavelet filter improves the adaptivity of wavelet transforms at the cost of slightly increasing the computational complexity of the forward pass for each graph filtering. However, the parameter complexity is significantly reduced especially for large graphs. 

\section{Experiments}\label{sec:exp}
In this section, we evaluate the proposed networks on two benchmark graph representation learning tasks: semi-supervised node classification and graph classification. Ablation studies are further conducted for a thorough analysis.

\subsection{Semi-supervised Node Classification}
The semi-supervised node classification aims to learn the representations of nodes to predict their labels. 
A small set of node labels are observed during training and the model is optimized by minimizing the cross-entropy loss between predictions and the observed labels.

\subsubsection{Datasets and Baselines} \emph{Cora}, \emph{Citeseer} and \emph{Pubmed} are three benchmark datasets for semi-supervised node classification task. Each dataset consists of a single graph where documents are represented with nodes and citation links are indicated by edges. Each node is given with features corresponding to the bag-of-words representation of the document and a label indicating the document category. The details of the datasets are presented in Table~\ref{table1}. Following the common experimental setting of prevailing models~\cite{kipf2016semi,xu2018graph}, for each dataset, 20 labeled nodes of each class are adopted for training, 500 nodes for validation, and 1000 nodes for testing.

We evaluate the proposed network against traditional approaches as well as state-of-the-art spatial and spectral GNNs. The \emph{traditional approaches} include label propagation (LP)~\cite{zhu2003semi}, semi-supervised embedding (SemiEmb)~\cite{weston2012deep}, manifold regularization (ManiReg)~\cite{belkin2006manifold}, graph embeddings (DeepWalk)~\cite{perozzi2014deepwalk}, iterative classification algorithm (ICA)~\cite{Lu:2003:LC:3041838.3041901} and Planetoid~\cite{yang2016revisiting}. The \emph{spatial models} include MoNet~\cite{monti2017geometric},  GraphSAGE~\cite{hamilton2017inductive}, and GAT~\cite{velivckovic2017graph}. The \emph{spectral models} include Fourier-based SGNNs (e.g., Spectral CNN~\cite{bruna2013spectral}, ChebyNet~\cite{defferrard2016convolutional}, GCN~\cite{kipf2016semi}, CayleyNet~\cite{levie2018cayleynets}, GraphARMA~\cite{bianchi2021graph}) as well as wavelet-based SGNNs ( e.g., GWNN~\cite{xu2018graph}, HANet~\cite{li2020fast} and UFG~\cite{zheng2021how}). Besides, we also compare with advance graph neural networks GCNII~\cite{chen2020simple} and PDTNet~\cite{luo2021learning}. 

\subsubsection{Experimental Settings}
We implement our models with an NVIDIA 1080Ti GPU. For fair comparison, we follow the same model configurations as prevailing SGNNs~\cite{kipf2016semi,xu2018graph} with two lifting-based adaptive graph wavelet layers with the ReLU function in the first layer and Softmax in the last layer. The number of hidden units is set to 16. Threshold is set for the  diffusion wavelets to remove noises and improve the computational efficiency as in~\cite{xu2018graph}. Models are trained with Adam optimizer for a maximum of 1000 epochs with an initial learning rate of 0.02. The threshold $\theta$ for soft-thresholding is set to 0.001. The training is early stopped if the validation loss does not decrease for 100 consecutive epochs. Grid search strategy is adopted to select the optimal hyper-parameters, including the number of lifting blocks, the scale and threshold of diffusion wavelets, weight decay and dropout rate. Table~\ref{table2} presents the optimal hyper-parameters for various datasets. We conduct 10 independent runs with 10 random seeds.  

\begin{table}[!t]
\renewcommand{\baselinestretch}{1.0}
\renewcommand{\arraystretch}{1.0}
\setlength\tabcolsep{3.5pt}
\centering
\caption{The selected Hyper-parameters of the proposed models.}\label{table2}
\begin{tabular}{l|cccccc}
\hline\hline
Datasets& blocks& scale & threshold & weight decay& dropout rate\\
\hline
\emph{Cora}&1&0.7&1e{-6}&1e{-3}&0.8\\
\emph{Citeseer}&1&0.5&1e{-6}&1e{-3}&0.5\\
\emph{PubMed}&1&0.7&1e{-7}&1e{-3}&0.5\\
\hline\hline
\end{tabular}
\end{table}

The results of traditional approaches are retrieved from~\cite{yang2016revisiting} and spatial models (except for GraphSAGE~\cite{hamilton2017inductive} reported in~\cite{liao2019lanczos}) and those of other spectral models (except for GWNN~\cite{xu2018graph} and UFG~\cite{zheng2021how}) are taken from respective works. We evaluate the advanced graph neural networks with the codes released by the authors. To make a fair comparison, the number of hidden units of GWNN and UFG is set to 16 and other hyper-parameters (i.e., learning rate, weight decay, and dropout rate) are finetuned with grid search according to the spaces recommended in respective papers. For advanced graph networks, we also evaluate GCNII with 16 units to make fair comparisons with other spectral models, and we denote it as GCNII$_{16}$ in Table~\ref{table3}. The optimal hyper-parameters for GCNII are set as recommended in~\cite{chen2020simple} and those for both GCNII$_{16}$ and PDTNet are selected with grid search according to the recommended spaces (if available). We run these models 10 times with different random seeds.

\subsubsection{Classification Performance}
Table~\ref{table3} shows that the proposed models achieve superior or comparative performance with state-of-the-art SGNNs on the three datasets. Especially, we achieve best performance on \emph{Cora} and \emph{Pubmed}. We suspect the possible reasons for the degraded performance on \emph{Citeseer} are two folds. The first one is that \emph{Citeseer} contains some isolated nodes, which will prevent the information propagation of diffusion wavelets as well as the lifting process in our model. The second one is that its average node degree is relatively small, which will also hamper the information propagation in lifting. Notably, though the soft-thresholding filtering operation has no learnable parameters, we significantly outperform other wavelet-based SGNNs that are based on manually-designed wavelets and parameter-intensive filters (i.e., GWNN~\cite{xu2018graph}, HANet~\cite{li2020fast}, UFGConvS and UFGConvR~\cite{zheng2021how}) on almost all datasets, which validates the effectiveness of the proposed adaptive graph wavelet filters.

\begin{table}[!t]
\renewcommand{\baselinestretch}{1.0}
\renewcommand{\arraystretch}{1.0}
\setlength\tabcolsep{3.5pt}
\centering
\caption{Statistics of datasets for node classification.}\label{table1}
\begin{tabular}{l|cccccc}
\hline\hline
Datasets& Nodes& Edges&Avg Deg & Classes&Features& Label Rates\\
\hline
\emph{Cora}&2708&5429&3.90&7&1433&5.2\%\\
\emph{Citeseer}&3327&4732&2.77&6&3703&3.6\%\\
\emph{PubMed}&19717&44338&4.50&3&500&0.3\%\\
\hline\hline
\end{tabular}
\end{table}
\begin{table}[!t]
\renewcommand{\baselinestretch}{1.0}
\renewcommand{\arraystretch}{1.0}
\setlength{\tabcolsep}{12pt}
\centering
\caption{Semi-supervised node classification accuracy (\%) on \emph{Cora}, \emph{Citeseer}, and \emph{PubMed}. The top-3 results are highlighted in bold. ($^{\star}$ results are reproduced with the publicly released codes. UFGConvS and UFGConvR denote the models with soft-thresholding and ReLU function respectively.)}\label{table3}
\begin{tabular}{l|ccc}
\hline\hline
{Methods}&\emph{Cora}&\emph{Citeseer}&\emph{PubMed}\\
\hline
MLP&55.1&46.5&71.4 \\
ManiReg~\cite{belkin2006manifold}&59.5&60.1&70.7\\
SemiEmb~\cite{weston2012deep}&59.0&59.6&71.7\\
LP~\cite{zhu2003semi}&68.0&45.3&63.0\\
DeepWalk~\cite{perozzi2014deepwalk}&67.2&43.2&65.3\\
ICA~\cite{Lu:2003:LC:3041838.3041901}&75.1&69.1&73.9\\
Planetoid~\cite{yang2016revisiting}&75.7&64.7&77.2\\
\hline
MoNet~\cite{monti2017geometric}&81.7$\pm$0.5 &$-$ &78.8$\pm$0.4\\
GraphSAGE~\cite{hamilton2017inductive}&74.5$\pm$0.8 &67.2$\pm$1.0 &76.8$\pm$0.6\\
GAT~\cite{velivckovic2017graph}&83.0$\pm$0.7 &\bf{72.5$\pm$0.7} &79.0$\pm$0.3 \\
\hline
GCNII$^\star$~\cite{chen2020simple}& \bf{85.4$\pm$0.4}&\bf{73.1$\pm$0.7} &\bf{79.8$\pm$0.4} \\
GCNII$_{16}^\star$~\cite{chen2020simple}& \bf{83.6$\pm$1.4}&71.9$\pm$0.2 &78.9$\pm$0.5 \\
PDTNet$^\star$~\cite{luo2021learning}& 81.5$\pm$0.4& 71.6$\pm$0.3&78.6$\pm$0.7\\
\hline
Spectral CNN~\cite{bruna2013spectral}&73.3 &58.9 &73.9 \\
ChebyNet~\cite{defferrard2016convolutional}&81.2 &69.8 &74.4 \\
GCN~\cite{kipf2016semi}&81.5 &70.3 &79.0 \\
CayleyNet~\cite{levie2018cayleynets}&81.9$\pm$0.7&-&- \\
GraphARMA~\cite{bianchi2021graph}&83.4$\pm$0.6&\bf 72.5$\pm$0.4&78.9$\pm$0.3\\
\hline
HANet~\cite{li2020fast}&81.9 &70.1 & 79.3\\
GWNN$^\star$~\cite{xu2018graph}& 81.6$\pm$0.7 &70.5$\pm$0.6 &78.6$\pm$0.3\\
UFGConvS$^\star$~\cite{zheng2021how}&82.3$\pm$0.8 &71.2$\pm$0.7 &76.2$\pm$1.4\\
UFGConvR$^\star$~\cite{zheng2021how}&82.9$\pm$1.2 &\bf{72.3$\pm$0.7} &77.5$\pm$1.2\\
LGWNN& 83.4$\pm$0.6 &71.1$\pm$0.4 &\bf79.5$\pm$0.5 \\
LGWNN$_{wide}$&\bf 83.9$\pm$0.4 &71.9$\pm$0.6 & \bf{79.5$\pm$0.3}\\
\hline\hline
\end{tabular}
\end{table}

\begin{table*}[!t]
\renewcommand{\baselinestretch}{1.0}
\renewcommand{\arraystretch}{1.0}
\centering
\caption{Statistics of datasets for graph classification( $\#$ represents number)}\label{table4}
\begin{tabular}{l|c|c|c|c|c|c}
\hline\hline
    Datasets& \# of Graphs & \# of Classes & Average $\#$ of nodes& Average $\#$ of edges& Avg $\#$ of degree& $\#$ of node labels   \\
    \hline
    \emph{DD} & 1178  & 2& 284.32 &715.66& 5.03 &89   \\
    \emph{PROTEINS}  &1113   &2& 39.06& 72.82 & 3.73 & 3 \\
    \emph{NCI1} & 4110 &2 & 29.87 &32.30 &2.17 &37 \\
    \emph{NCI109} & 4127 &2 & 29.68 &32.13 & 2.17& 38\\
   \emph{Mutagenicity}&4337&2&30.32&30.77&2.07&14\\
   \hline\hline   
  \end{tabular}
\end{table*}

Regarding advanced graph neural networks, Table~\ref{table3} demonstrates that GCNII achieves state-of-the-art performance in semi-supervised node classification with deep models (i.e., 64, 32, and 16 layers are employed in GCNII and GCNII$_{16}$ on \emph{Cora}, \emph{Citeseer}, and \emph{Pubmed}, respectively), validating the effectiveness of the \emph{initial residual} and \emph{identity mapping} in alleviating the over-smoothing problem of vanilla GCNs. Compared with GCNII$_{16}$, with same width and much less layers (2 layers are employed in LGWNN on all the three datasets), LGWNN achieves comparable and superior performance on \emph{Cora} and \emph{Pubmed} respectively, indicating the effectiveness and efficiency of the proposed adaptive graph wavelet filters in extracting useful graph representations. The performance of LGWNN is inferior to that of GCNII$_{16}$ on \emph{Citeseer}, which may also be explained by the isolated nodes and small average node degree of \emph{Citeseer}. Table~\ref{table3} also reveals that the width of graph filtering layers in GCNII contributes to the performance gain. For example, by reducing the width of GCNII to 16 as other spectral models, the performance on all the three datasets is consistently degraded. Inspired by this observation, we further increase the width of the proposed model to explore its potential. We select the width from $\{32,64\}$ and denote the model as LGWNN$_{wide}$. LGWNN$_{wide}$ yields improved classification accuracy on \emph{Cora} and \emph{Citeseer} and outperforms GCNII$_{16}$ on \emph{Cora} and \emph{Pubmed}. These facts indicate that both width and depth influence the capacity of graph neural networks and also demonstrate the superiority of the proposed adaptive graph wavelet networks. Although PDTNet improves the robustness (as verified in~\cite{luo2021learning}), the performance gain over vanilla GCNs is not evident in our experiments and LGWNNs significantly outperform PDTNet on all the three datasets.

\begin{table}[!t]
\renewcommand{\baselinestretch}{1.0}
\renewcommand{\arraystretch}{1.0}
\setlength\tabcolsep{3.5pt}
\centering
\caption{The selected Hyper-parameters of the proposed models.}\label{table6}
\begin{tabular}{l|cccc}
\hline\hline
Datasets& blocks& scale & threshold0 &threshold1\\
\hline
\emph{DD}&1&1.0&0.001&0.01\\
\emph{PROTEINS}&1&0.7&0.01&0.01\\
\emph{NCI1}&1&1.0&0.01&0.1\\
\emph{NCI109}&1&1.0&0.01&0.01\\
\emph{Mutagenicity}&1&1.0&0.01&0.1\\
\hline\hline
\end{tabular}
\end{table}
\begin{table*}[!t]
\renewcommand{\baselinestretch}{1.0}
\renewcommand{\arraystretch}{1.0}
\centering
\caption{Graph classification accuracy (\%) on \emph{PROTEINS}, \emph{NCI1}, \emph{NCI109}, \emph{Mutagenicity}, and \emph{DD}. The best three results are in bold.}\label{table5}
\begin{tabular}{l|l|ccccc}
\hline\hline
\multicolumn{2}{c|}{Models}&\emph{PROTEINS}&\emph{NCI1}&\emph{NCI109}&\emph{Mutagenicity}&\emph{DD}\\
\hline
\multirow{3}{*}{Spatial models}&GraphSAGE~\cite{hamilton2017inductive}&69.91$\pm$6.55&75.69$\pm$1.34	&73.14$\pm$2.12&80.37$\pm$1.26&77.09$\pm$2.67\\
\cline{2-7}&GAT~\cite{velivckovic2017graph}&\bf72.23$\pm$4.59&78.64$\pm$1.25&76.15$\pm$1.84&81.73$\pm$2.53&76.84$\pm$3.83\\
\cline{2-7}&GIN~\cite{xu2018how}&70.44$\pm$5.36&\bf79.00$\pm$1.54&\bf79.47$\pm$2.37&82.30$\pm$1.47&76.67$\pm$2.26\\
\hline
\multirow{3}{*}{Fourier-based models}&GCN~\cite{kipf2016semi}&70.36$\pm$5.52&75.11$\pm$1.72&76.59$\pm$1.78&80.89$\pm$1.92&\bf78.38$\pm$5.20\\
\cline{2-7}&ChebyNet~\cite{defferrard2016convolutional}&72.14$\pm$6.13&\bf79.29$\pm$1.23&\bf78.71$\pm$2.18&\bf{82.71$\pm$1.62}&\bf78.37$\pm$3.73\\
\cline{2-7}&ARMA~\cite{bianchi2021graph}&71.61$\pm$5.61&\bf{79.88$\pm$1.67}&\bf{80.02$\pm$2.28}&\bf82.63$\pm$1.62&76.92$\pm$4.67\\
\hline
\multirow{3}{*}{Wavelet-based models}&GWNN~\cite{xu2018graph}&\bf73.35$\pm$3.71&69.03$\pm$1.82&69.79$\pm$1.67&74.26$\pm$2.29&75.04$\pm$4.55\\
\cline{2-7}&LGWNN&\bf{74.02$\pm$5.23}&78.97$\pm$2.07&	76.37$\pm$1.75&	\bf82.47$\pm$1.90&	\bf{78.72$\pm$4.33}\\
\hline\hline
\end{tabular}
\end{table*}

\subsection{Graph Classification}
Given a graph $\mathcal G$ with $N$ node features $\mathbf{X}\in \mathbb{R}^{N\times f_{in}}$ and adjacency matrix $\mathbf{W}\in \mathbb{R}^{N\times N}$, graph classification aims to learn a graph-level representation $h(\mathcal G)\in \mathbb{R}^{f_{out}}$ from the pair $(\mathbf{X},\mathbf{W})$ to predict the label for the graph $\mathcal{G}$. 
\subsubsection{Datasets and Baselines} We evaluate the proposed model and state-of-the-art GNNs on five benchmark bioinformatics graph datasets, namely, \emph{PROTEINS}, \emph{NCI1}, \emph{NCI109}, \emph{DD} and \emph{Mutagenicity}. All the datasets contain more than $1000$ graphs with varying graph structures (e.g., average number of nodes and edges, average degree of node) and node attributes. The detailed statistics are presented in Table~\ref{table4}. 

Following the common practice of the graph classification~\cite{bianchi2021graph}, the node features are the concatenation of the one-hot encoding of node labels, node degrees, and clustering coefficients. We split each dataset into 10 folds. 8 of them are adopted for training and the remaining 2 folds are used for validation and testing, respectively.

In addition to state-of-the-art GNNs (i.e., ARMA~\cite{bianchi2021graph} and GIN~\cite{xu2018how}) specifically deigned for graph classification, we also extend prevailing GNNs to graph classification based on Pytorch Geometric library~\cite{fey2019fast}, including spectral models (i.e, GCN~\cite{kipf2016semi}, ChebNet~\cite{defferrard2016convolutional}, GWNN~\cite{xu2018graph}) and spatial models (i.e, GraphSAGE~\cite{hamilton2017inductive} and GAT~\cite{velivckovic2017graph}). 
\subsubsection{Experimental Settings}
For fair comparison, all the models adopt the same architecture which consists of three graph convolutional layers, a global mean pooling layer, and a fully-connected layer as GraphARMA~\cite{bianchi2021graph}. The features learned in the three layers are concatenated for multi-scale graph representations before being fed into the global pooling layer. All the models are trained with Adam optimizer with learning rate of 0.001 for a maximum of 1000 epoch. The learning is terminated if the validation loss does not decrease in consecutive 50 epochs. The batch size and the feature dimensionality of the graph convolutional layers for all models are set to 32. Dropout rate is fixed as 0.5. For the proposed models, the number of lifting blocks, the scale and threshold for wavelets (threshold0), and the threshold for soft-thresholding (threshold1) are selected from \{1,2,3\}, \{0.5,0.7,1.0\}, \{0.01,0.001,0.0001\}, and \{0.01,0.1\} via grid search. The optimal configurations are presented in Table~\ref{table6}. All the models are independently initialized and run 10 times.

\subsubsection{Classification Performance}
Table~\ref{table5} reports the mean results and standard deviations. The spectral-based graph neural networks generally outperform spatial-based models and achieve state-of-the-art performance on all the five datasets. It confirms that spectral-based models are better to capture the graph structural information. The proposed model achieves comparative results with state-of-the-art spectral graph neural networks i.e., GraphARMA~\cite{bianchi2021graph} which has much more parameters than our model (approximately 8$\times$). In particular, we achieve the best performance on \emph{PROTEINS} and \emph{DD} datasets as they have relatively large average node degree (i.e., 3.73 and 5.03, respectively) compared to the others, which facilitates adaptive wavelets learning in lifting. Notably, we significantly outperform previous wavelet-based SGNNs (i.e., GWNN) by a large margin on almost all datasets with 9.94\% on \emph{NCI1}, 6.85\% on \emph{NCI109}, 8.21\% on \emph{Mutagenicity}, and 3.68\% on \emph{DD}. These facts validate the superiority of the proposed lifting-based adaptive graph wavelet filters against the non-adaptive ones. The performance of the proposed model on \emph{NCI1} and \emph{NCI109} is slightly below the state-of-the-art spectral models. We suspect the reasons are the excessive sparse graph (i.e., relative small average node degree) and complex node features (i.e., high-dimensional node features), which increases the difficulty in modeling the signal correlations in lifting.

\begin{figure}[!t]
\renewcommand{\baselinestretch}{1.0}
\centering
\includegraphics[width=0.8\columnwidth]{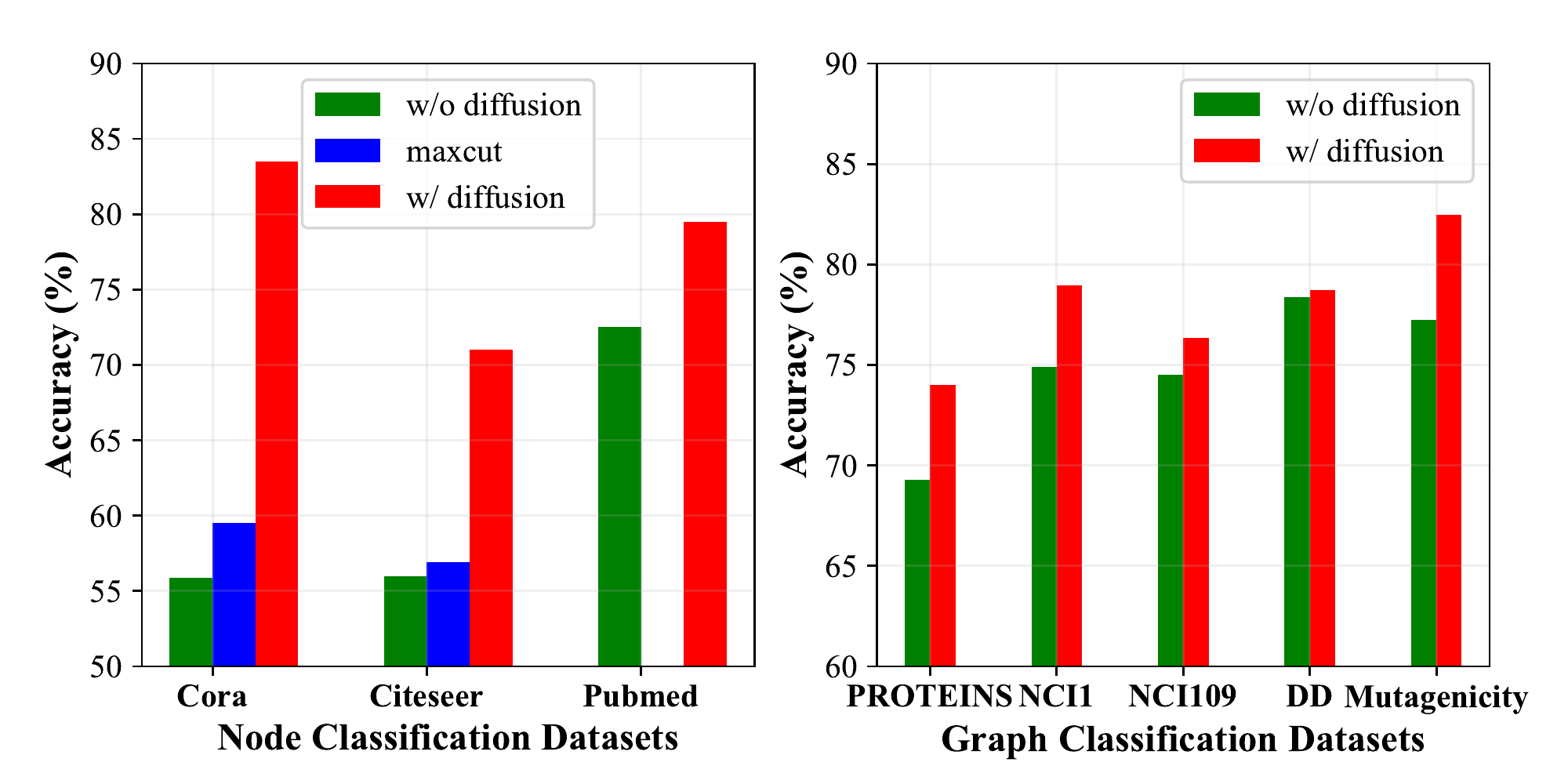}
\caption{The performance of LGWNN without diffusion wavelet transforms, with Maxcut scheme, and with diffusion wavelet transforms.}\label{fig5}
\end{figure}

\subsection{Ablation Studies}
Ablation studies are further conducted on both node and graph classification tasks to validate our design choices, 
to study the influence of important hyper-parameters, and to compare the model complexity.
\subsubsection{Effectiveness of Lifting Based on Diffusion Wavelets}
We first study the effectiveness of lifting based on diffusion wavelets by removing the diffusion wavelet transform from LGWNN and replacing it with Maxcut scheme (i.e., maximally preserving the edges before lifting). We adopt the popular Kernighan–Lin algorithm~\cite{kernighan1970efficient} to cut graphs. However, its heavy computational complexity restricts its applicability to datasets with large or varying-size graphs. So we only apply it to \emph{Cora} and \emph{Citeseer}.
\figurename~\ref{fig5} shows that the performance is consistently degraded on all the datasets after removing the diffusion wavelets. Moreover, we outperform the models with Maxcut scheme on \emph{Cora} and \emph{Citeseer} by a large margin. These facts confirm the effectiveness of our scheme in relieving the \emph{structural information loss}.

\begin{figure}[!t]
\renewcommand{\baselinestretch}{1.0}
\centering    \includegraphics[width=0.8\columnwidth]{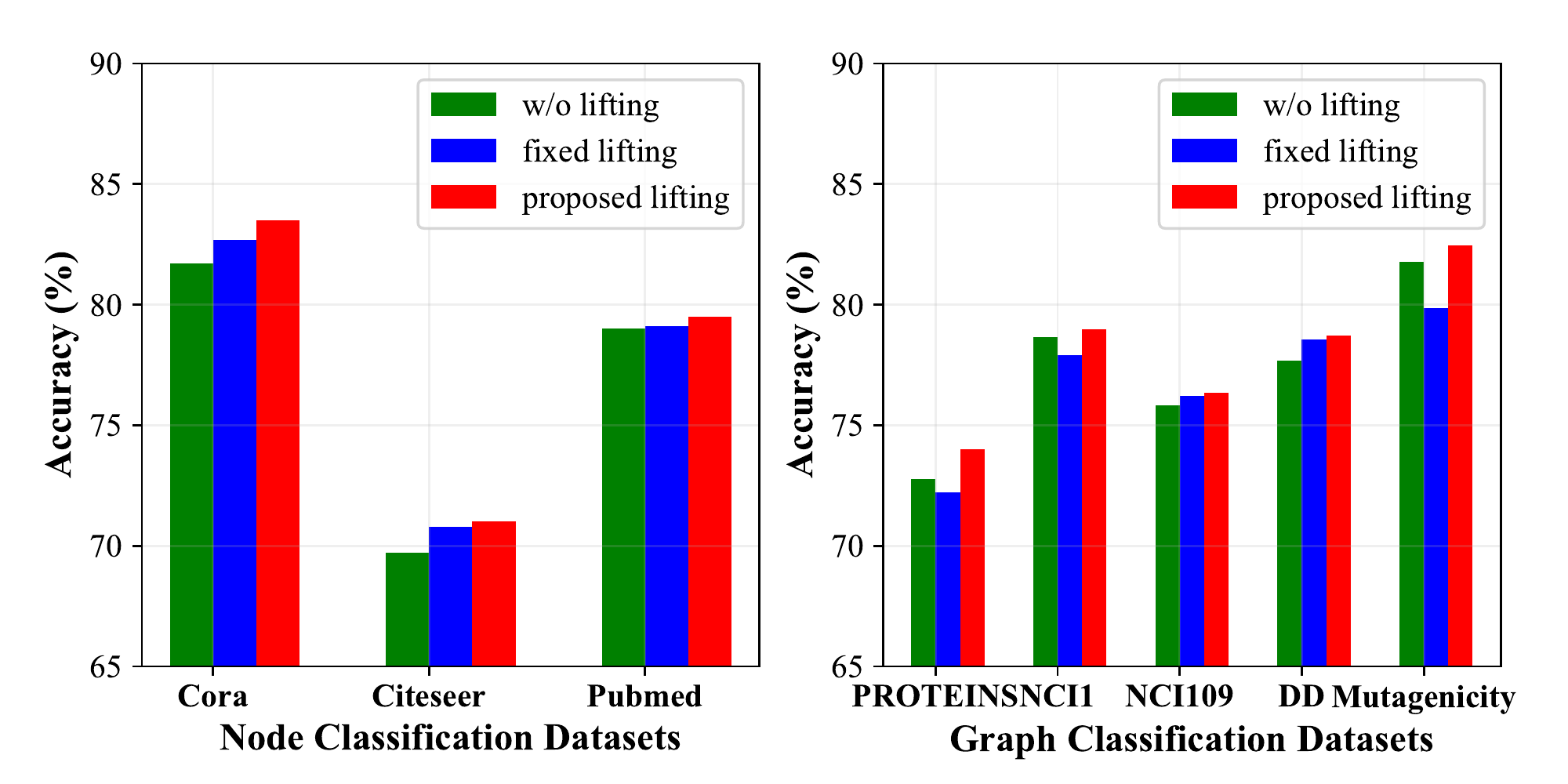}
\caption{The performance of LGWNN without lifting, with fixed lifting, and with the proposed lifting.}\label{fig6}
\end{figure}
\begin{figure}[!t]
\renewcommand{\baselinestretch}{1.0}
\centering
\subfigure{\includegraphics[width=0.38\columnwidth]{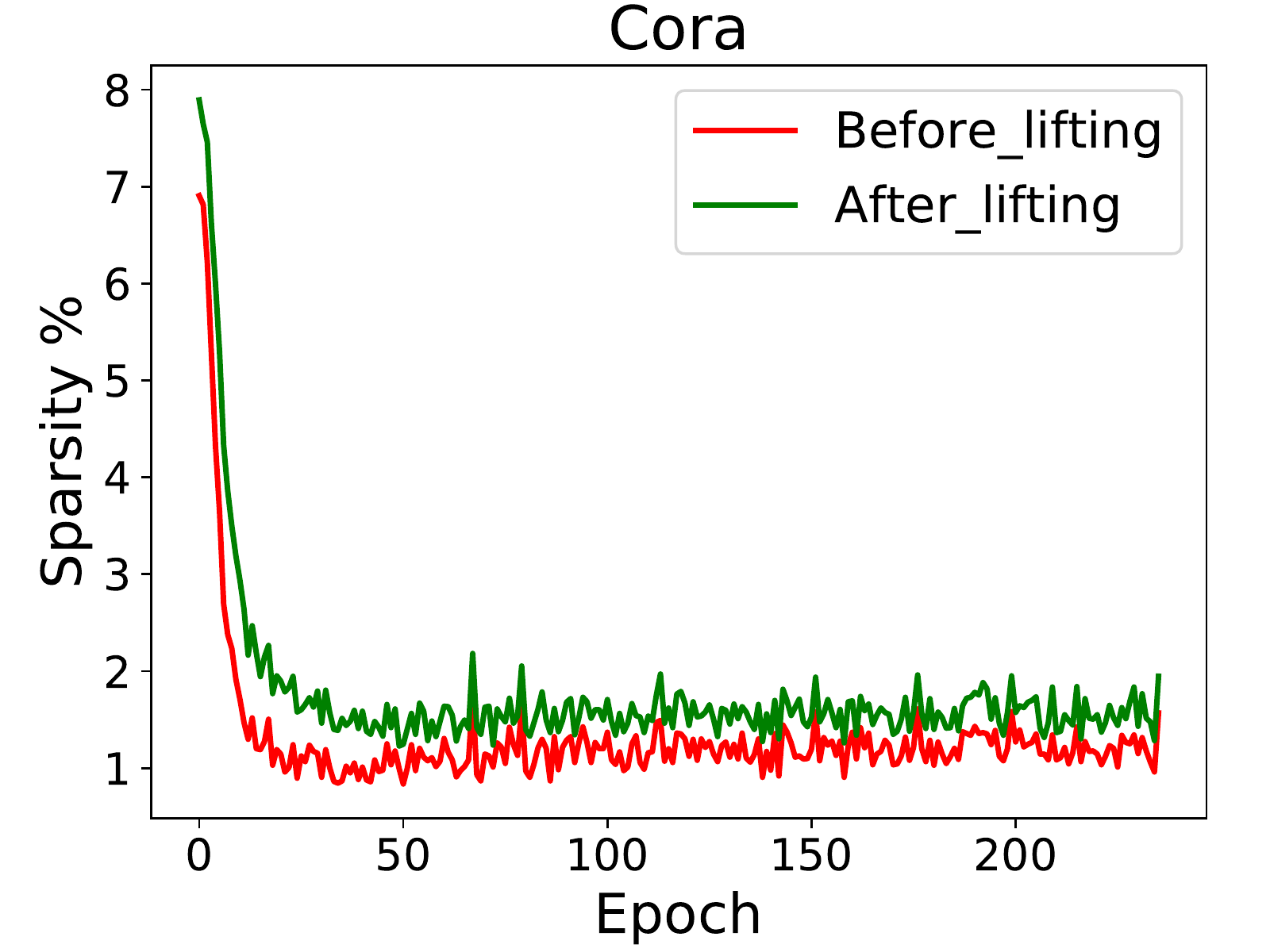}}\subfigure{\includegraphics[width=0.38\columnwidth]{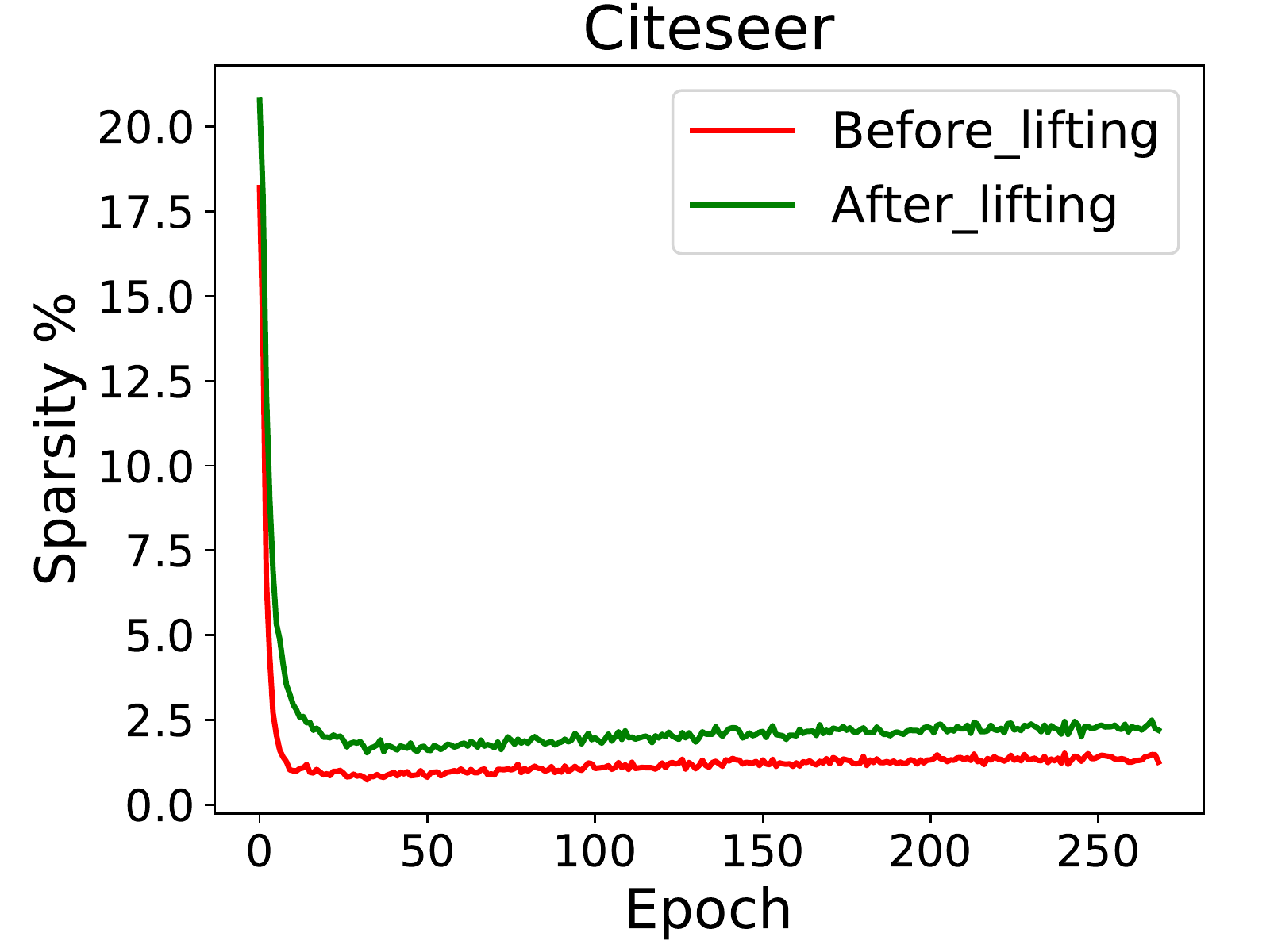}}
\caption{The effectiveness of the proposed lifting structures in improving wavelet coefficients sparsity on \emph{Cora} and \emph{Citeseer} during training.}\label{fig7}
\end{figure}
\begin{figure}[!t]
\renewcommand{\baselinestretch}{1.0}
\centering
\includegraphics[width=0.8\columnwidth]{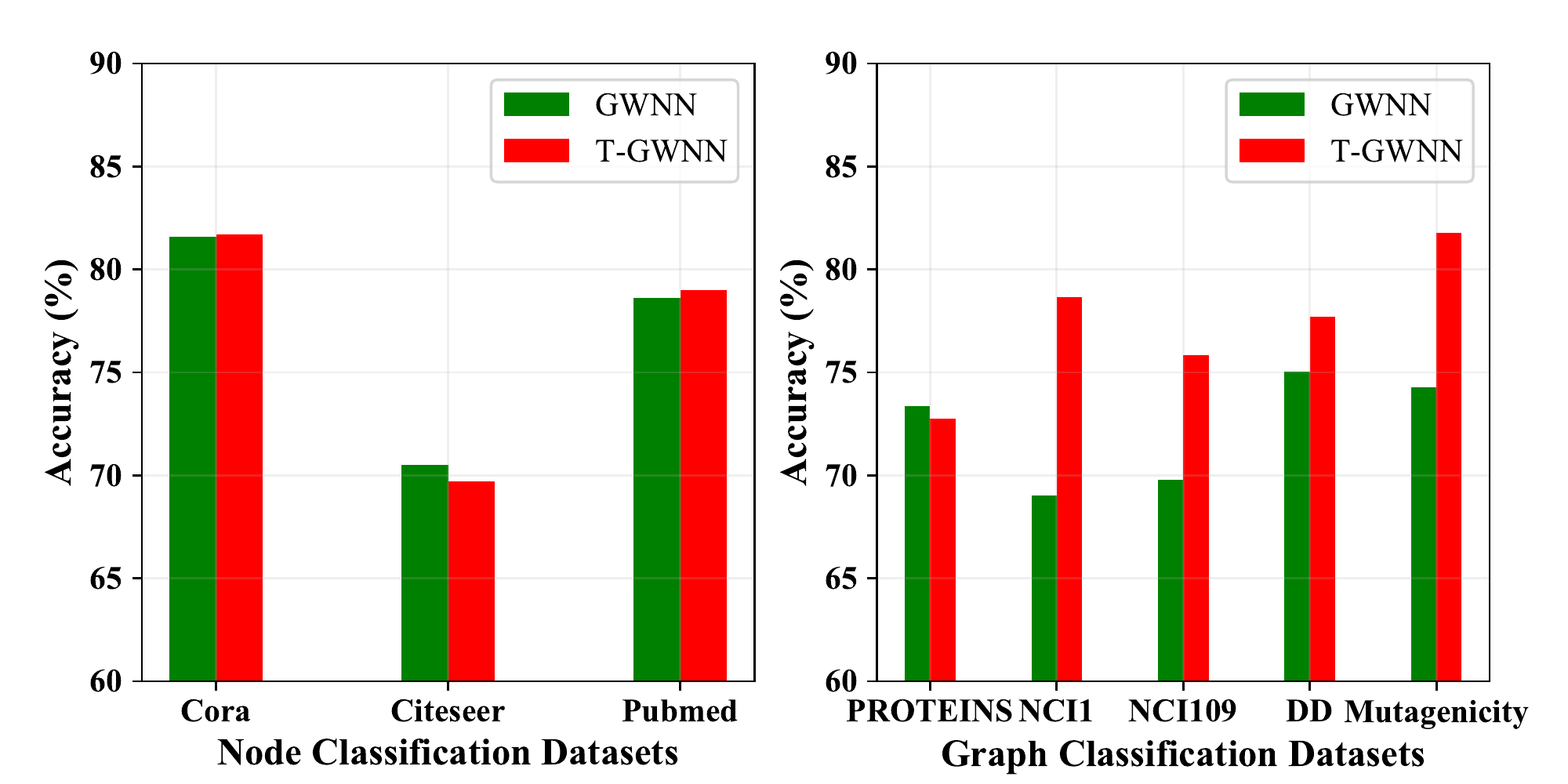}
\caption{The performance of models with learnable spectral filters (GWNN) and with the soft-thresholding filtering operation (T-GWNN) on different datasets.}\label{fig8}
\end{figure}

\subsubsection{Effectiveness of the Proposed Lifting Operations}
We further evaluate the effectiveness of the proposed learnable lifting structures by dropping the lifting structures and replacing it with the fixed lifting operations (i.e., non-learnable)~\cite{narang2009lifting} in LGWNN. Note that the lifting structure with fixed operations constructs 5/3-like wavelets, which satisfies the \emph{1-order vanishing moment condition} but could not adapt to graph signals and tasks at hand. \figurename~\ref{fig6} shows that the performance of models with the proposed lifting structures consistently outperforms those models without lifting and with fixed lifting operations. 

\begin{figure}[!t]
\renewcommand{\baselinestretch}{1.0}
\centering
\includegraphics[width=0.8\columnwidth]{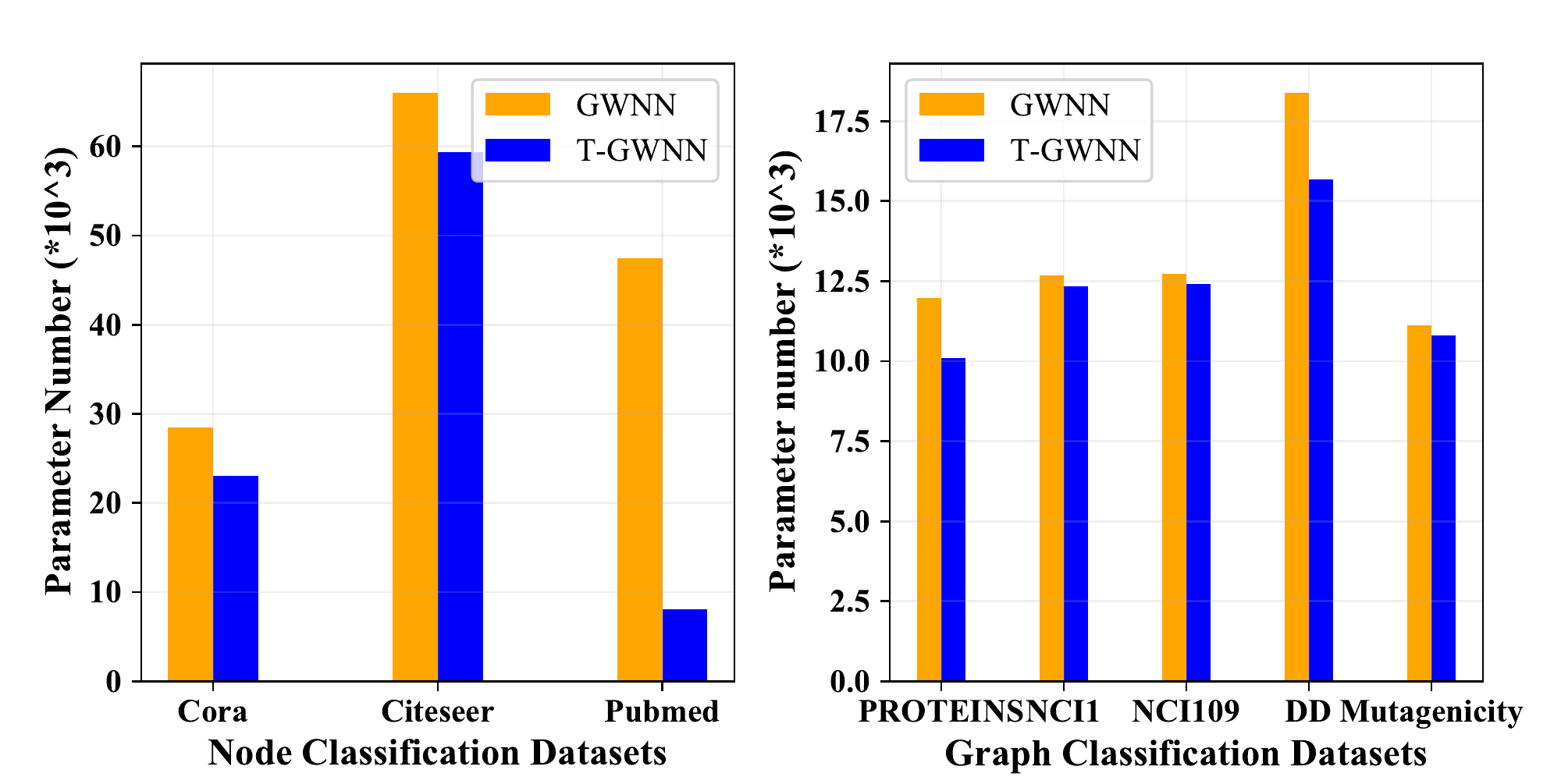}
\caption{The number of parameter of models with learnable spectral filters (GWNN) and with the soft-thresholding filtering operation (T-GWNN) on different datasets.}\label{fig9}
\end{figure}
\begin{figure}[!t]
\renewcommand{\baselinestretch}{1.0}
\centering
\subfigure{\includegraphics[width=0.48\columnwidth]{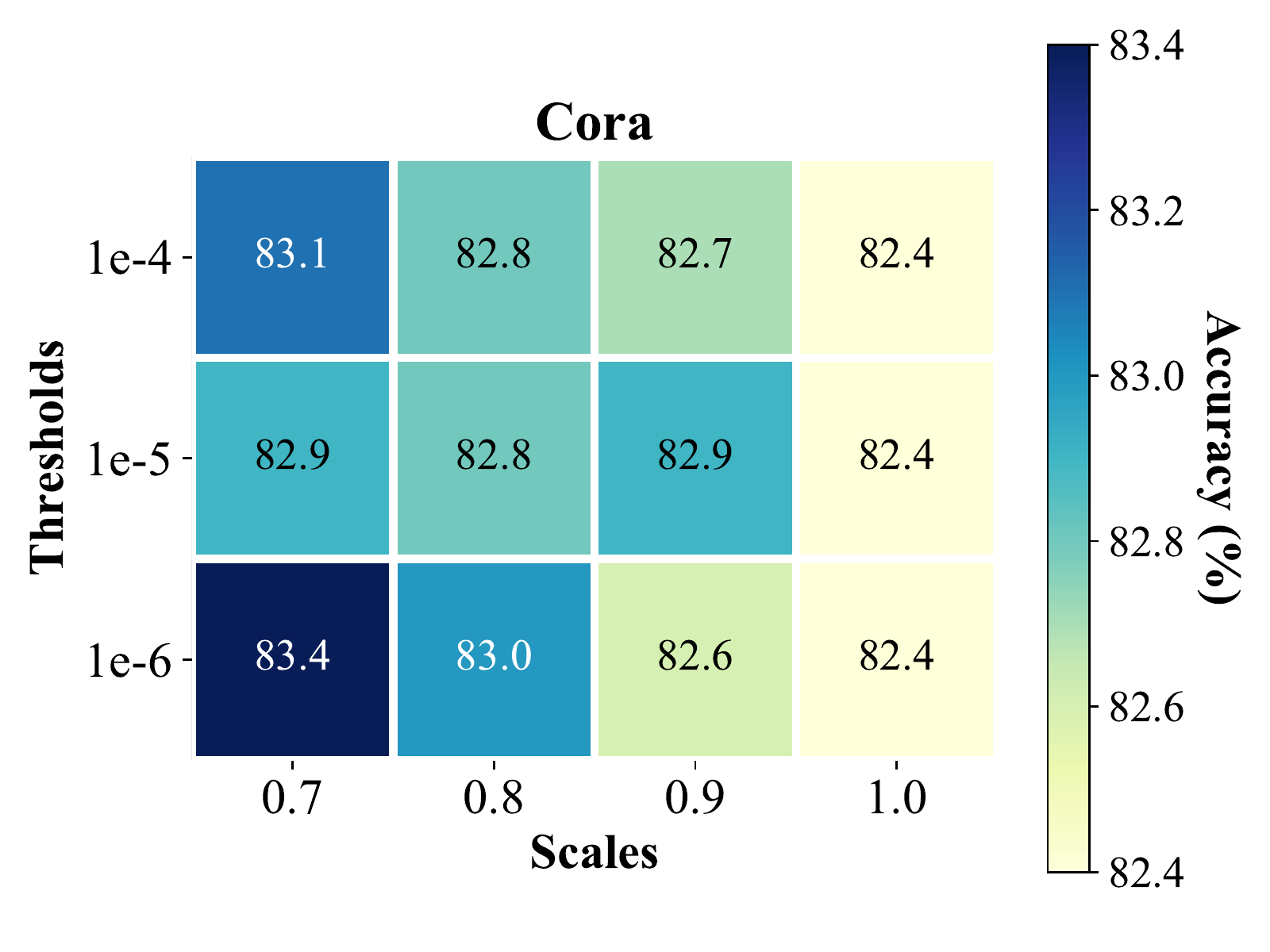}}\subfigure{\includegraphics[width=0.48\columnwidth]{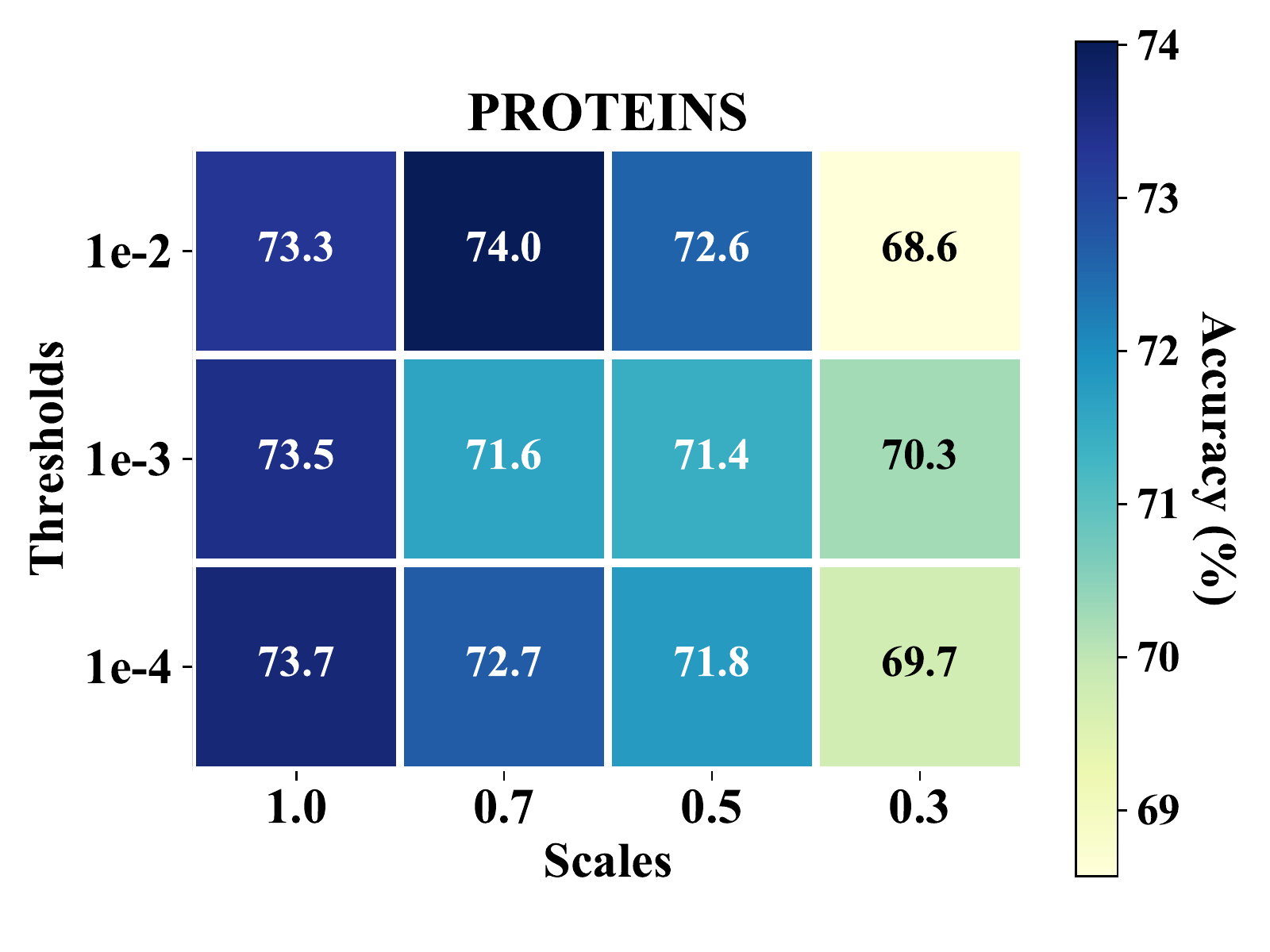}}
\caption{The influence of the scale and threshold of diffusion wavelets.}\label{fig10}
\end{figure}
\begin{figure}[!t]
\renewcommand{\baselinestretch}{1.0}
\centering
\subfigure[\emph{Cora}]{\includegraphics[width=0.45\columnwidth]{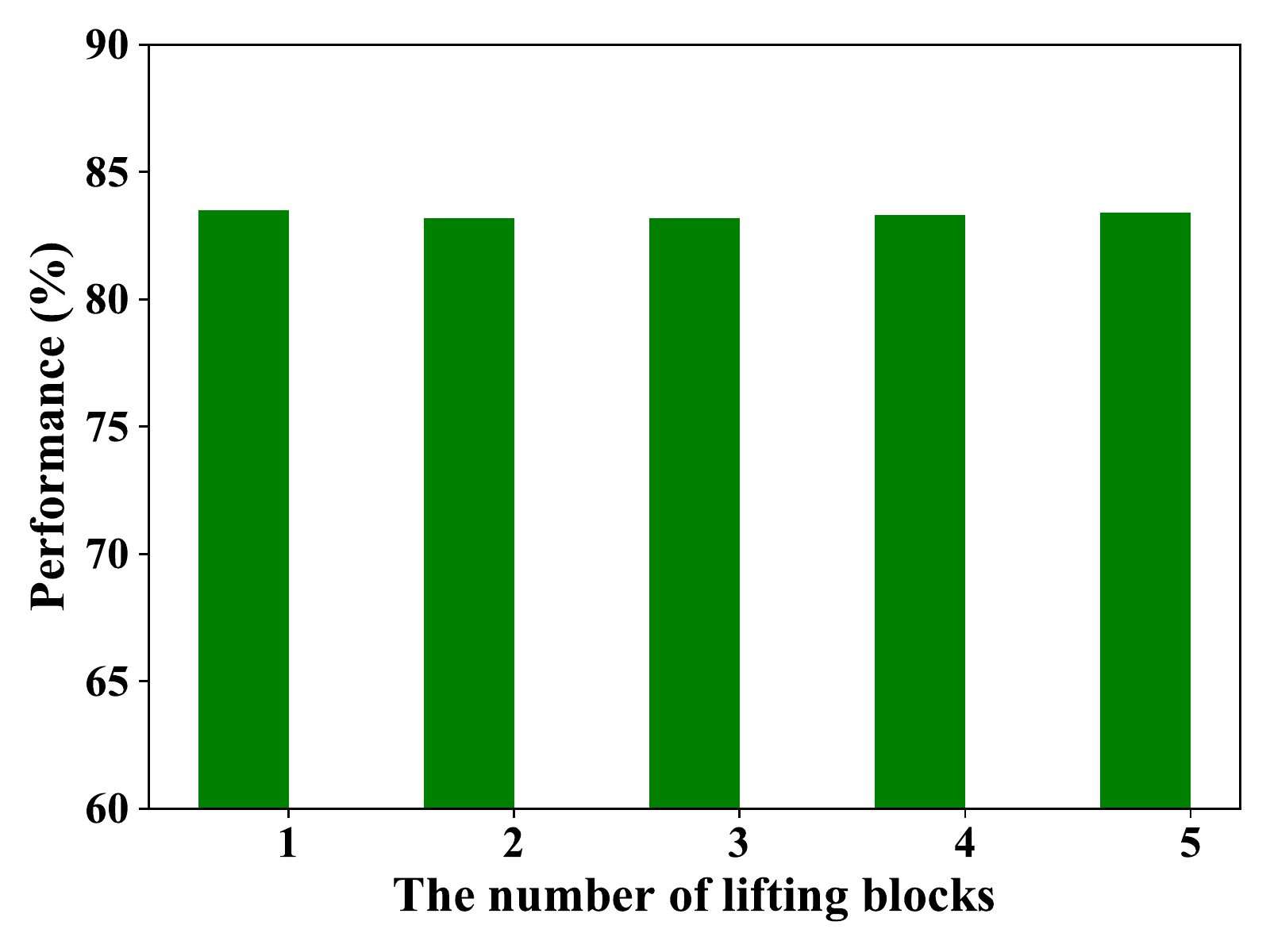}}\subfigure[\emph{PROTEINS}]{\includegraphics[width=0.45\columnwidth]{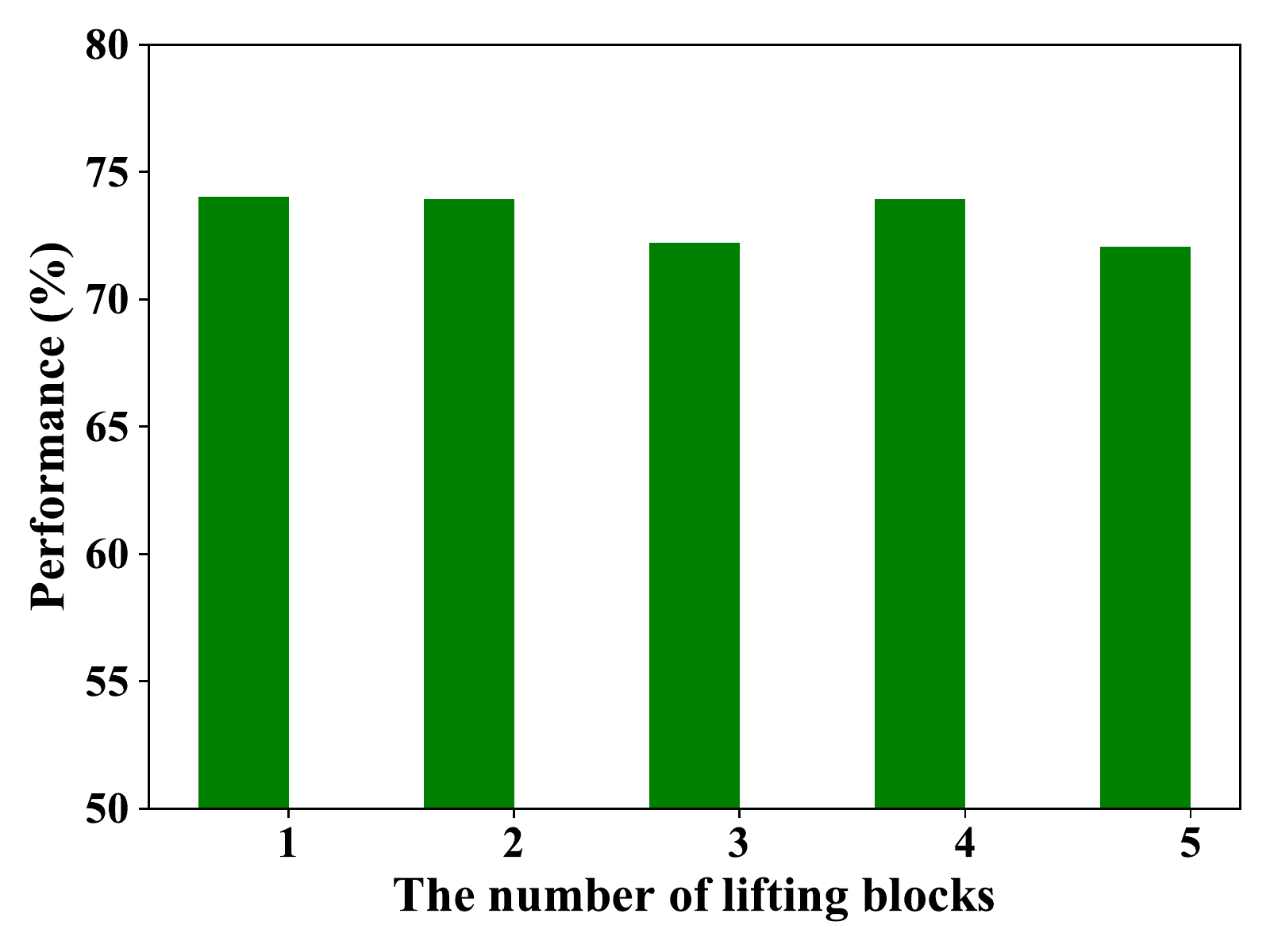}}
\caption{The influence of the number of lifting blocks.}\label{fig11}
\end{figure}
\begin{figure*}[!t]
\renewcommand{\baselinestretch}{1.0}
\centering
\includegraphics[width=1.8\columnwidth,height=0.4\columnwidth]{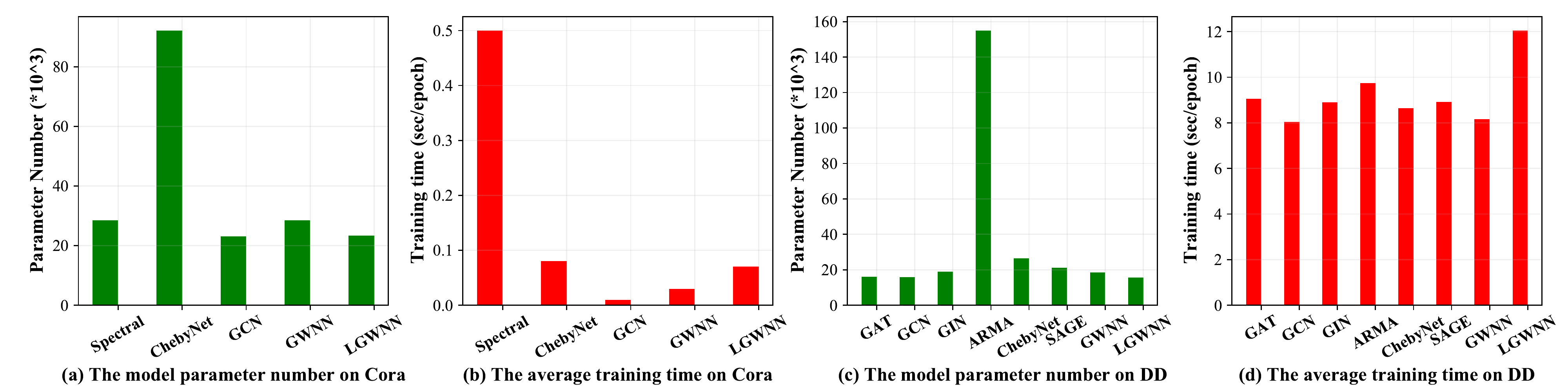}
\caption{The comparison of complexity of different models on different datasets. The metric for computational complexity is the average training time per epoch and for parameter complexity, it is the number of learnable parameters.}\label{fig12}
\end{figure*}

To show that the proposed lifting structure is effective in exploiting the node correlations to produce more small wavelet coefficients, we further compare the wavelet coefficients sparsity (i.e., the ratio of wavelet coefficients below the given threshold) before and after lifting in the first filtering layers on \emph{Cora} and \emph{Citeseer}. \figurename~\ref{fig7} shows that the wavelet coefficients sparsity is consistently improved after lifting.  
\subsubsection{Effectiveness of Soft-thresholding Filtering}

To show the superiority and applicability of the soft-thresholding filtering scheme, we perform experiments on the pioneering wavelet-based model i.e., GWNN~\cite{xu2018graph}. We replace its parameter-intensive multiplication operator with soft-thresholding filtering operation, leading to T-GWNN. The thresholds for soft-thresholding operations are set to 0.001 for node classification and 0.01 for graph classification. 

The results and parameter complexity of GWNN and T-GWNN are presented in \figurename~\ref{fig8} and \figurename~\ref{fig9}, respectively. \figurename~\ref{fig8} shows that, for node classification task, comparable or even superior performance can be achieved by T-GWNN. 
It also shows that, for graph classification where datasets typically consist of varying-size graphs, the performance of T-GWNN significantly outperforms that of GWNN on most of the datasets except \emph{PROTEINS}. The reason is that most of the datasets consists of graphs with highly-varying sizes, which poses great challenge for the multiplication operator learning while \emph{PROTEINS} contains graphs with similar size. \figurename~\ref{fig9} shows that the parameter number is consistently reduced on all datasets with soft-thresholding filtering. These facts demonstrate the superiority of the soft-thresholding filtering and may advocate the study of learning sparse graph representation.

\subsubsection{Influence of Hyper-parameters}
We also explore the effects of important hyper-parameters including the scale and threshold of diffusion wavelets and the number of lifting blocks in graph filters on \emph{Cora} and \emph{PROTEINS}. 
 
\figurename~\ref{fig10} shows that the performance is quite stable on both datasets. We can also observe that the performance is generally more stable and better in some scales (e.g., 0.7 and 0.8 for \emph{Cora} and 1.0 and 0.7 for \emph{PROTEINS}). Too large scales will remove too much high-frequency information from the input graph signals while too small scales would preserve too much noisy high frequency signals which will also degrade the performance. A general principle for selecting these two parameters is left for future study. 

We further evaluate the influence of the number of lifting blocks. The number of lifting blocks is selected from \{1,2,3,4,5\}. \figurename~\ref{fig11} shows that the performance is stable with increasing number of lifting blocks and even slightly degraded. Best performance on both datasets is achieved with one lifting block. This shows that the proposed structure-aware attention-based lifting operations are effective in capturing signal correlations, leading to desirable graph wavelets.  

\subsubsection{Complexity Analysis}
In this part, we compare the average training time (sec/epoch) and parameter complexity across different models on \emph{Cora} and \emph{DD} datasets. All the models for node classification are evaluated on a NVIDIA 1080 Ti GPU and for graph classification are studied on a NVIDIA 2080 Ti GPU. The model configurations and experimental settings are kept the same for fair comparison. 

\figurename~\ref{fig12}(a) shows that LGWNN has less parameters than Spectral CNN, GWNN and ChebyNet, similar to that of GCN on \emph{Cora}. \figurename~\ref{fig12}(b) shows that the training speed of LGWNN is comparable to ChebyNet, and is slightly slower compared to GWNN due to the additional lifting structures. 

Regarding graph classification tasks, \figurename~\ref{fig12}(c) shows that the parameter complexity of LGWNN is much less than GraphARMA~\cite{bianchi2021graph} and similar to the others. \figurename~\ref{fig12}(d) shows that we need slightly more training time than other models as the cost for adaptive graph wavelets.

\section{Conclusion}\label{sec:con}
In this paper, we proposed a novel class of lifting-based adaptive graph wavelet networks that implements graph filtering with adaptive graph wavelet transforms. Efficient and scalable lifting structures with structure-aware attention-based prediction and update operations are proposed to efficiently learn wavelets adapted to graph signals and tasks at hand. The locality, sparsity, and vanishing moments are guaranteed by design.  We further proposed a soft-thresholding operation for wavelet filtering, resulting in efficient, scalable, and interpretable graph wavelet filters. Moreover, permutation-invariant and feature transformation layers are further adopted to facilitate permutation-invariant graph representation learning. Experiments on both node- and graph-level tasks demonstrate the effectiveness and efficiency of the proposed model. In the future, we will apply the proposed model in graph signal reconstruction and denoising tasks.

\ifCLASSOPTIONcaptionsoff
  \newpage
\fi

\bibliographystyle{IEEEtran}
\bibliography{IEEEabrv,Reference.bib}
\end{document}